\documentclass[11pt]{article}

\usepackage[a4paper,margin=2.5cm]{geometry}
\usepackage{amsmath,amssymb,amsthm,mathtools}
\usepackage{microtype}
\usepackage{booktabs,tabularx,array}
\usepackage{enumitem}
\usepackage{xcolor}
\usepackage{graphicx}
\usepackage{caption}
\usepackage{subcaption}
\usepackage[section]{placeins}
\newcolumntype{L}[1]{>{\raggedright\arraybackslash}p{#1}}
\newcolumntype{Y}{>{\raggedright\arraybackslash}X}
\usepackage{tikz}
\usetikzlibrary{positioning,arrows.meta,calc,fit,backgrounds,decorations.pathreplacing,shapes.geometric,shapes.misc}
\usepackage[colorlinks=true,linkcolor=blue!50!black,citecolor=green!40!black,urlcolor=blue!50!black]{hyperref}
\usepackage[capitalise,noabbrev]{cleveref}

\setlist{itemsep=2pt,topsep=4pt}

\theoremstyle{definition}
\newtheorem{property}{Property}
\crefname{property}{Property}{Properties}
\theoremstyle{plain}
\newtheorem{theorem}{Theorem}[section]
\newtheorem{proposition}[theorem]{Proposition}
\theoremstyle{definition}
\newtheorem{definition}[theorem]{Definition}

\newcommand{\R}{\mathbb{R}}
\newcommand{\Z}{\mathbb{Z}}

\newcommand{\M}{\mathcal{M}}
\newcommand{\T}{\mathcal{T}}
\newcommand{\Pset}{\mathcal{P}}
\newcommand{\tr}{\operatorname{tr}}
\newcommand{\rank}{\operatorname{rank}}
\newcommand{\diag}{\operatorname{diag}}
\newcommand{\Pareto}{\operatorname{Pareto}}
\newcommand{\What}{\widehat{W}}
\newcommand{\bits}{\operatorname{bits}}

\tikzset{
  inode/.style={circle, draw, minimum size=6.5mm, inner sep=0pt, fill=blue!8, font=\scriptsize},
  znode/.style={circle, draw, minimum size=6.5mm, inner sep=0pt, fill=black!25, font=\scriptsize},
  leaf/.style={rectangle, draw, minimum width=6.5mm, minimum height=5mm, inner sep=1pt, fill=orange!14, font=\scriptsize},
  zleaf/.style={rectangle, draw, minimum width=6.5mm, minimum height=5mm, inner sep=1pt, fill=black!20, font=\scriptsize},
  opnode/.style={circle, draw, thick, minimum size=6.5mm, inner sep=0pt, fill=green!12, font=\small},
  shared/.style={rectangle, rounded corners, draw, thick, fill=purple!10, font=\scriptsize, inner sep=3pt},
  boxn/.style={rectangle, rounded corners, draw, fill=yellow!10, font=\scriptsize, inner sep=3pt},
  lab/.style={font=\scriptsize\itshape, text=black!70},
  shc/.style={draw, fill=blue!12, rounded corners=1.5pt},
  prc/.style={draw, fill=orange!16, rounded corners=1.5pt},
  zec/.style={draw, fill=black!18, rounded corners=1.5pt},
  emc/.style={draw=black!35, dashed, rounded corners=1.5pt},
}
\newcommand{\icell}[6]{%
  \draw[#5] ({#1+0.78*(#2-1)+0.04},{-0.6*(#4)}) rectangle ({#1+0.78*(#3)-0.04},{-0.6*(#4)+0.5})
    node[midway, font=\tiny] {#6};}

\title{\textbf{Every Weight Is a Path of Shared Discrete Decisions}}
\author{
  Ammar Ali\thanks{Corresponding author: \texttt{ammarali32@itmo.ru}}\\
  {\small LEAD Lab, ITMO University}
  \and
  Stamatis Lefkimmiatis\\
  {\small LEAD Lab, ITMO University}
}
\date{October 2026}

\begin{document}
\maketitle

\begin{abstract}
We introduce TWIG (Tree-structured Weights with Inherited Gates), a representation in which every weight of a neural network is the end of a path of small discrete decisions: a sign, a shift by a power of two, a zero gate, an index into a shared table. Weights that belong together share the upper part of their paths, and each decision is inherited by every weight below it, so it is stored once and applied once. This simple change has three consequences. Any trained network can be written as a TWIG, and when the decisions are powers of two its weight layers become circuits made almost entirely of wires and adders, suited to chips whose weights are fixed at fabrication. Pruning, quantization, low-rank factorization, codebooks and weight sharing become edits of one object, so a single optimizer can choose among all of them for a whole network. And choosing the decisions
is the ground-state problem of a higher-order Boltzmann machine whose interactions follow the paths of the tree, the problem class that annealers and quantum optimizers are built to solve. Today these questions are handled apart: each compression method brings its own format and solver, a circuit with fixed weights needs a dedicated constant
multiplier for every weight, and discrete solvers reach only narrow subproblems such as rounding.
Weight formats have long been moving toward shared decisions, through group scales, shared exponents, microexponents and sparse weights; TWIG takes this trend to its end, and every common format is a shallow TWIG at no extra cost. On five weight matrices of a large language model, a flat tree whose signed power-of-two terms are chosen per weight for circuit cost, with no group scales, needs 9--15\% fewer additions than GPTQ at equal layer error at 3- and 4-bit accuracy on typical matrices, and 20--55\% fewer on matrices with strong outliers. The properties of TWIG are proved; the measurements we report are first evidence on selected models and layers.
\end{abstract}
\section{Introduction}\label{sec:intro}

Modern neural networks are trained in the continuum. Gradient descent needs a smooth landscape to follow, so weights are modelled as real numbers and learning is described as motion through a high-dimensional real space; much of what we understand about optimization and generalization rests on this view. Deployment, however, takes place in a discrete world. On digital hardware, every weight is a finite bit pattern drawn from a fixed format, every product is computed by a circuit of logic gates, and every byte has to be moved through a memory hierarchy. The passage from the continuous model that was learned to the discrete object that is executed is therefore unavoidable, and it has grown into a research field of its own.

That field has grown in pieces. Pruning removes weights, quantization restricts them to finite grids, low-rank factorization replaces matrices by products of thinner ones, codebook methods map blocks of weights to entries of a shared table, and weight sharing reuses parameters across layers. Each family brings its own storage format, its own solver and often its own kernel, and combining them (four bits in one layer, half of the weights pruned in another, a low-rank correction in a third) remains largely a matter of manual design, without a common yardstick. Two further questions sit almost entirely outside this picture. The first is the cost of \emph{computing} with the weights rather than storing them: in digital logic a multiplier is considerably larger and more energy-hungry than an adder \cite{horowitz2014energy}, and in circuits whose weights are fixed at fabrication, such as printed classifiers \cite{mubarik2020printed} or chips that hardwire an entire language model \cite{taalas2026hc1}, every weight calls for a constant multiplier of its own. The second is the reach of discrete optimization: choosing discrete weights is a combinatorial problem, yet annealers, samplers and quantum optimizers, which are built for exactly such problems, have so far been applied only to narrow parts of it, such as the rounding of a single layer \cite{nagel2020adaround}. What is missing, we argue, is a representation of weights that is discrete by construction, that makes the structure shared by weights explicit, and that exposes the cost of applying them.

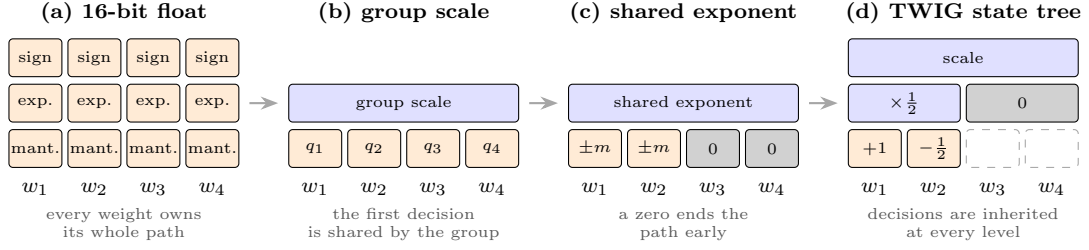
\begin{figure}[t]
\centering
\begin{tikzpicture}[x=1cm,y=1cm]
\foreach \k in {1,...,4} {
  \icell{0}{\k}{\k}{0}{prc}{sign}
  \icell{0}{\k}{\k}{1}{prc}{exp.}
  \icell{0}{\k}{\k}{2}{prc}{mant.}}
\icell{3.7}{1}{4}{1}{shc}{group scale}
\foreach \k in {1,...,4} {\icell{3.7}{\k}{\k}{2}{prc}{$q_{\k}$}}
\icell{7.4}{1}{4}{1}{shc}{shared exponent}
\icell{7.4}{1}{1}{2}{prc}{$\pm m$}
\icell{7.4}{2}{2}{2}{prc}{$\pm m$}
\icell{7.4}{3}{3}{2}{zec}{$0$}
\icell{7.4}{4}{4}{2}{zec}{$0$}
\icell{11.1}{1}{4}{0}{shc}{scale}
\icell{11.1}{1}{2}{1}{shc}{$\times\frac12$}
\icell{11.1}{3}{4}{1}{zec}{$0$}
\icell{11.1}{1}{1}{2}{prc}{$+1$}
\icell{11.1}{2}{2}{2}{prc}{$-\frac12$}
\icell{11.1}{3}{3}{2}{emc}{}
\icell{11.1}{4}{4}{2}{emc}{}
\foreach \X in {0,3.7,7.4,11.1} {
  \foreach \k in {1,...,4} {\node[font=\scriptsize] at ({\X+0.78*\k-0.39},-1.48) {$w_{\k}$};}}
\node[font=\scriptsize\bfseries] at (1.56,0.85) {(a) 16-bit float};
\node[font=\scriptsize\bfseries] at (5.26,0.85) {(b) group scale};
\node[font=\scriptsize\bfseries] at (8.96,0.85) {(c) shared exponent};
\node[font=\scriptsize\bfseries] at (12.66,0.85) {(d) TWIG state tree};
\node[font=\tiny, text=black!65, align=center] at (1.56,-1.95) {every weight owns\\its whole path};
\node[font=\tiny, text=black!65, align=center] at (5.26,-1.95) {the first decision\\is shared by the group};
\node[font=\tiny, text=black!65, align=center] at (8.96,-1.95) {a zero ends the\\path early};
\node[font=\tiny, text=black!65, align=center] at (12.66,-1.95) {decisions are inherited\\at every level};
\foreach \X in {3.12,6.82,10.52} {\draw[-{Stealth}, thick, black!35] (\X+0.1,-0.35) -- (\X+0.48,-0.35);}
\end{tikzpicture}
\caption{From formats in use to TWIG. Each column of boxes is the description of one weight, read from top to bottom; a box that spans several weights is a decision they share, stored once.
(a)~A 16-bit float stores a sign, an exponent and a mantissa for every weight. 
(b)~Group quantization shares one scale among the weights of a group. 
(c)~Block formats share an exponent,and a zero ends a weight's description. 
(d)~A state tree shares decisions at every level, and each decision is inherited by every weight below it: one scale for all four weights, a halving for the first pair, an ``off'' switch for the second pair, after which nothing more is stored (dashed). Blue: shared decisions; orange: private decisions; grey: zero.}
\label{fig:hook}
\end{figure}

The formats used in practice have been moving toward such a representation for some time (\cref{fig:hook}). Group quantization lets the weights of a group share one scale \cite{frantar2023gptq,lin2024awq}; block floating point and the microscaling formats let a block of weights share one exponent \cite{rouhani2020msfp,rouhani2023mx}; sparse formats let the description of a weight end as soon as it is known to be zero; and codebook methods describe a whole block of weights by one or a few entries of a shared table
\cite{egiazarian2024aqlm,tseng2024quipsharp}. Each of these steps records a decision once for many weights instead of once for every weight, and each has paid off in memory, bandwidth and speed. A few formats already stack two or three levels of shared scales, such as shared microexponents \cite{rouhani2023sharedmx} and HiFloat4 \cite{luo2026hif4}. Yet each format shares one fixed kind of decision, at levels fixed in advance, in a format of its own.

We take this trend to its natural end. Our proposal can be stated in a single sentence:
\begin{quote}
\itshape Write every weight as the end of a path of small discrete decisions, and let weights that belong together share the upper part of the path.
\end{quote}
The result is a tree over the entries of a weight matrix (\cref{fig:hook}d). Each internal node holds one small decision (a shift by a power of two, a sign, a zero gate, an index into a shared table), which is inherited by every weight below it; each leaf holds the little that remains private to a single weight; and the value of a weight is the product of the decisions along its path. We call this representation TWIG, for \emph{Tree-structured Weights with Inherited Gates}, and refer to an individual tree as a \emph{state tree}. A state tree still describes an ordinary real matrix, so the learned model and its deployed form remain directly comparable; what the tree adds is an explicit account of how the matrix is stored, which decisions its entries share, and what it costs to apply.

This single change addresses the three questions above at once, because the same tree can be read in three ways (\cref{tab:readings}). Read as a \emph{compressed model}, a state tree contains today's formats as shallow special cases, and pruning, quantization, factorization, codebooks and sharing all become edits of the same object; methods developed separately become comparable and combinable, and a single budgeted optimizer can choose among all of them for a whole network (\cref{sec:compress}). Read as a \emph{circuit}, a tree whose decisions are powers of two multiplies by shifting, which in fixed hardware is merely a way of routing wires; its weight layers become circuits made almost entirely of wires and adders, and precision and circuit size become a single quantity (\cref{sec:circuits}). Read as an \emph{optimization problem}, choosing the decisions of a tree is the search for the lowest-energy state of a higher-order Boltzmann machine whose couplings are supplied by the data and whose higher-order terms follow the paths of the tree, the problem class that classical samplers, annealers and quantum optimizers are built to solve (\cref{sec:bm}). Finally, because the decisions are explicit, a network could be trained directly in this discrete space, so that the trained object is already the deployed one; we outline this direction in \cref{sec:learning}.

\paragraph{Contributions.}
\begin{itemize}[leftmargin=*]
\item We introduce TWIG, a representation of neural network weights as trees of shared discrete decisions, together with its two costs: storage bits and circuit additions (\cref{sec:trees}).
\item We establish its basic properties: it contains the number formats in use today at no extra cost, it loses nothing in the worst case and uses fewer bits where weights share structure, and any matrix can be approximated, to any precision, by a sum of power-of-two trees that is applied with shifts and additions alone (\cref{sec:express}).
\item We show that the main families of compression methods are edits of one object, chosen by a single budgeted optimizer that is exact under additive costs (\cref{sec:compress}).
\item We show how a state tree maps onto a circuit of wires and adders, and report a first measurement on five weight matrices of a large language model: a flat tree whose signed power-of-two terms are chosen per weight saves 9--15\% of the additions of GPTQ at equal layer error at 3- and 4-bit accuracy on typical matrices, and 20--55\% on matrices with strong outliers (\cref{sec:circuits}).
\item We show that choosing the states of a tree is the ground-state problem of a higher-order Boltzmann machine whose higher-order terms follow the paths of the tree, which connects weight selection to classical and quantum discrete optimizers
(\cref{sec:bm}).
\end{itemize}
Several of the properties restate classical results in the language of trees, as marked in \cref{app:formal}; the results specific to state trees are the reach of shared shifts, the equivalence of two encodings, the exact optimizer with shared subtrees and the higher-order Boltzmann machine. \Cref{sec:directions} collects first evidence on real models and beyond them, and \cref{sec:related} relates TWIG to number formats, decision diagrams and multiplier-free networks. Precise definitions, statements and proofs are collected in \cref{app:formal}.
\section{From numbers to state trees}\label{sec:trees}

\subsection{A number is a path of decisions}

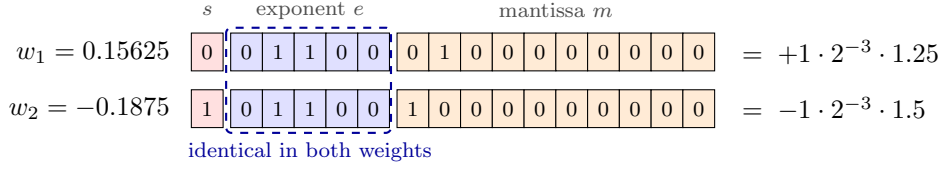
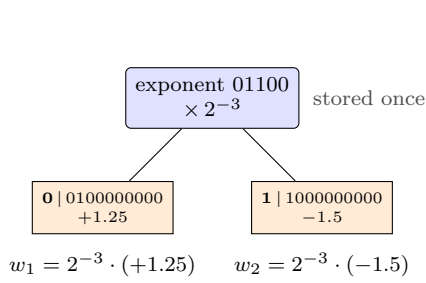
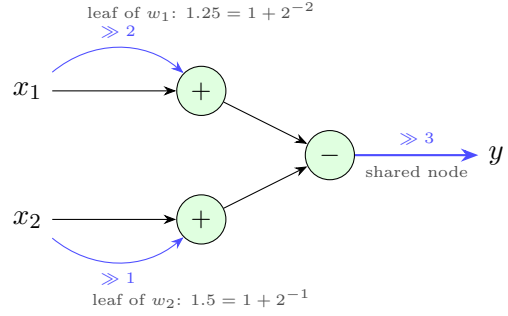
\begin{figure}[t]
\centering
\tikzset{
  sbit/.style={draw, fill=red!12, minimum width=4.2mm, minimum height=5mm, inner sep=0pt, font=\scriptsize},
  ebit/.style={draw, fill=blue!12, minimum width=4.2mm, minimum height=5mm, inner sep=0pt, font=\scriptsize},
  mbit/.style={draw, fill=orange!16, minimum width=4.2mm, minimum height=5mm, inner sep=0pt, font=\scriptsize},
  addr/.style={circle, draw, fill=green!12, minimum size=6.5mm, inner sep=0pt, font=\small},
}
\def\bitrow#1#2#3#4{%
  \node[sbit] at (2.81,#1) {#2};
  \foreach \b [count=\k from 0] in {#3} {\node[ebit] at ({3.33+0.42*\k},#1) {\b};}
  \foreach \b [count=\k from 0] in {#4} {\node[mbit] at ({5.53+0.42*\k},#1) {\b};}}
\begin{subfigure}[b]{\textwidth}
\centering
\begin{tikzpicture}
\node[font=\scriptsize, text=black!70] at (2.81,0.55) {$s$};
\node[font=\scriptsize, text=black!70] at (4.17,0.55) {exponent $e$};
\node[font=\scriptsize, text=black!70] at (7.42,0.55) {mantissa $m$};
\bitrow{0}{0}{0,1,1,0,0}{0,1,0,0,0,0,0,0,0,0}
\bitrow{-0.75}{1}{0,1,1,0,0}{1,0,0,0,0,0,0,0,0,0}
\node[anchor=east, font=\small] at (2.45,0) {$w_1=0.15625$};
\node[anchor=east, font=\small] at (2.45,-0.75) {$w_2=-0.1875$};
\node[anchor=west, font=\small] at (9.75,0) {$=\;+1\cdot2^{-3}\cdot1.25$};
\node[anchor=west, font=\small] at (9.75,-0.75) {$=\;-1\cdot2^{-3}\cdot1.5$};
\draw[blue!60!black, thick, dashed, rounded corners=2pt] (3.06,0.32) rectangle (5.24,-1.07);
\node[font=\scriptsize, blue!60!black] at (4.17,-1.32) {identical in both weights};
\end{tikzpicture}
\caption{Two weights in FP16: each stores its own sign, exponent and mantissa (16 bits each).}
\end{subfigure}

\vspace{3mm}
\begin{subfigure}[b]{0.44\textwidth}
\centering
\begin{tikzpicture}
\node[draw, rounded corners=2pt, fill=blue!12, font=\scriptsize, align=center] (n) at (0,0)
  {exponent $01100$\\$\times\,2^{-3}$};
\node[draw, fill=orange!16, font=\tiny, align=center] (l1) at (-1.45,-1.45) {$\mathbf{0}\,|\,0100000000$\\$+1.25$};
\node[draw, fill=orange!16, font=\tiny, align=center] (l2) at (1.45,-1.45) {$\mathbf{1}\,|\,1000000000$\\$-1.5$};
\draw (n) -- (l1); \draw (n) -- (l2);
\node[font=\scriptsize] at (-1.45,-2.2) {$w_1=2^{-3}\cdot(+1.25)$};
\node[font=\scriptsize] at (1.45,-2.2) {$w_2=2^{-3}\cdot(-1.5)$};
\node[font=\scriptsize, text=black!70, anchor=west] at (1.2,0) {stored once};
\end{tikzpicture}
\caption{The shared exponent is stored once; each leaf keeps its sign and mantissa. 27 bits
instead of 32, with exactly the same values.}
\end{subfigure}\hfill
\begin{subfigure}[b]{0.53\textwidth}
\centering
\begin{tikzpicture}[>=Stealth]
\node (x1) at (0,0) {$x_1$};
\node (x2) at (0,-1.7) {$x_2$};
\node[addr] (a1) at (2.3,0) {$+$};
\node[addr] (a2) at (2.3,-1.7) {$+$};
\draw[->] (x1) -- (a1);
\draw[->, blue!70] (x1.north east) to[bend left=40] node[above, font=\tiny] {$\gg2$} (a1.north west);
\draw[->] (x2) -- (a2);
\draw[->, blue!70] (x2.south east) to[bend right=40] node[below, font=\tiny] {$\gg1$} (a2.south west);
\node[addr] (s) at (4.0,-0.85) {$-$};
\draw[->] (a1) -- (s); \draw[->] (a2) -- (s);
\node (y) at (6.2,-0.85) {$y$};
\draw[->, blue!70, thick] (s) -- node[above, font=\tiny] {$\gg3$} node[below, font=\tiny, text=black!70] {shared node} (y);
\node[font=\tiny, text=black!70] at (2.3,1.05) {leaf of $w_1$: $1.25=1+2^{-2}$};
\node[font=\tiny, text=black!70] at (2.3,-2.75) {leaf of $w_2$: $1.5=1+2^{-1}$};
\end{tikzpicture}
\caption{Computing $y=w_1x_1+w_2x_2$ from the tree: three additions, no multiplication.}
\end{subfigure}
\caption{A first example. (a)~Two neighbouring weights stored in 16-bit floating point; their exponents are identical, yet the format stores them twice. (b)~Because a weight is the \emph{product} of its decisions, the shared exponent can be stored once in a node above both weights; each weight is recovered exactly as the product along its path. (c)~The same tree computes $w_1x_1+w_2x_2$ without forming either weight: each leaf becomes shifted copies of its own input ($\gg k$ shifts by $k$ binary positions, i.e.\ multiplies by $2^{-k}$), and the shared exponent is applied once, to the sum, as a shift by three positions. Three additions replace two multiplications and one addition.}
\label{fig:example}
\end{figure}

Consider a weight stored in 16-bit floating point, $w_1=0.15625$ (\cref{fig:example}a). Its sixteen bits are not read as one number but as three decisions: a sign bit $s$, a five-bit exponent $e$ and a ten-bit mantissa $m$. On the normal range the value they encode is a product of three factors, one per decision,
\[
  w\;=\;(-1)^{s}\cdot 2^{\,e-15}\cdot\Big(1+\frac{m}{1024}\Big),
\]
so that $w_1=(+1)\cdot2^{-3}\cdot1.25$. Read from left to right, the decisions form a path: the sign selects a half-line, the exponent an octave within it, and the mantissa a point within the octave. Now consider a second weight from the same matrix, $w_2=-0.1875=(-1)\cdot2^{-3}\cdot1.5$. Its sign and mantissa differ from those of $w_1$, but its exponent is the same: both weights lie in the same octave of magnitudes. In a standard format each weight owns its entire path, so this common exponent is stored twice.

\subsection{Sharing a decision at the same precision}

Because the value of a weight is a product of its decisions, a factor that two weights have in common can be taken out of both and stored once, in a node placed above them (\cref{fig:example}b). The node holds the exponent, $2^{-3}$; the two leaves below it hold what remains private to each weight, a sign and a mantissa. Each weight is recovered as the product of the decisions along its path from the node to its leaf, so both values are reproduced exactly, at the same precision as before, with $5+2\times11=27$ bits instead of $32$. The saving grows with the size of the group: $g$ weights that share an exponent need $5+11g$ bits instead of $16g$, a reduction that approaches $31\%$ for large groups without changing a single value. The weights of a trained layer rarely share an exponent exactly, but neighbouring weights tend to lie within a few octaves of one another. A second kind of shared decision covers this case: a node that holds a single bit and either leaves everything below it unchanged or halves it. Placed at several levels, such bits let weights of different but nearby magnitudes share most of their path, and the private part left at each leaf can then be shortened where accuracy allows. This is the general state tree of \cref{sec:trees-general}.

\subsection{Multiplying without rebuilding the weight}

The same structure changes how the weights are used. A layer never needs the weights themselves, only their products with its inputs, here $y=w_1x_1+w_2x_2$. Written along the tree,
\[
  y\;=\;2^{-3}\Big[\big(1+2^{-2}\big)\,x_1\;-\;\big(1+2^{-1}\big)\,x_2\Big].
\]
Each leaf acts on its own input, and the shared node acts once, on the sum of its subtree (\cref{fig:example}c). Every factor is a power of two, so every multiplication is a shift: $x_1+(x_1\gg2)$ and $x_2+(x_2\gg1)$ take one addition each, a subtraction combines them, and the shared exponent becomes a shift of the result by three positions, which in a circuit with fixed weights is only a way of routing wires. Three additions replace two multiplications and one addition, and neither weight is ever formed as a number. In general a mantissa has more nonzero digits than in this example, and each costs one addition; \cref{sec:express} shows that a few signed power-of-two terms per weight suffice for any desired precision, and \cref{sec:circuits} measures how many are needed on real weights.

\subsection{State trees in general}\label{sec:trees-general}

\begin{figure}[t]
\centering
\begin{tikzpicture}[
  level 1/.style={sibling distance=46mm},
  level 2/.style={sibling distance=23mm},
  level 3/.style={sibling distance=11.5mm},
  level distance=12mm]
\node[inode] (r) {$0$}
  child {node[inode] (a) {$0$}
    child {node[inode] (aa) {$1$}
      child {node[leaf] (l1) {$+1$}}
      child {node[leaf] (l2) {$0$}}}
    child {node[inode] (ab) {$0$}
      child {node[leaf] (l3) {$-\frac12$}}
      child {node[leaf] (l4) {$+1$}}}}
  child {node[inode] (b) {$1$}
    child {node[inode] (ba) {$0$}
      child {node[leaf] (l5) {$-1$}}
      child {node[leaf] (l6) {$+\frac12$}}}
    child {node[inode] (bb) {$1$}
      child {node[leaf] (l7) {$0$}}
      child {node[leaf] (l8) {$+1$}}}};
\node[boxn, above=5mm of r] (beta) {column base $\beta_j$};
\draw[-{Stealth}] (beta) -- (r);
\foreach \i in {1,...,8} {\node[font=\scriptsize, below=1.5mm of l\i] {$\rho_{\i}$};}
\begin{scope}[on background layer]
\draw[line width=2.2pt, red!60] (r.center) -- (b.center) -- (ba.center) -- (l6.center);
\end{scope}
\node[lab, anchor=west, align=left] at ($(bb.east)+(0.9,0.75)$) {%
  node bit $b\in\{0,1\}$: multiply everything\\
  below the node by $(\frac12)^{b}$;\\
  leaf state $a\in\{0,\pm1,\pm\frac12\}$.};
\node[anchor=north west, align=left, font=\small] at ($(l8.south east)+(0.35,0.15)$) {%
  $\What_{6j}=\rho_6\,\beta_j\cdot 1\cdot\tfrac12\cdot 1\cdot\big(+\tfrac12\big)$\\[1pt]
  $\phantom{\What_{6j}}=\tfrac14\,\rho_6\,\beta_j$};
\node[lab] at ($(l1.south west)+(-0.9,-0.25)$) {rows};
\end{tikzpicture}
\caption{The running example: a deeper state tree for one column $j$ of a matrix with eight rows. Each internal node holds one bit that halves everything below it or leaves it unchanged; each leaf holds a sign and a magnitude, possibly zero. The value of an entry is the product of the states on its path (red), times a row scale and a column base, \cref{eq:value}. Apart from $\rho_i\beta_j$, every factor is a power of two or zero. The column stores 7 node bits (6 if the root bit is folded into $\beta_j$) and 8 five-state leaves, about 3.2 bits per weight.}
\label{fig:anatomy}
\end{figure}
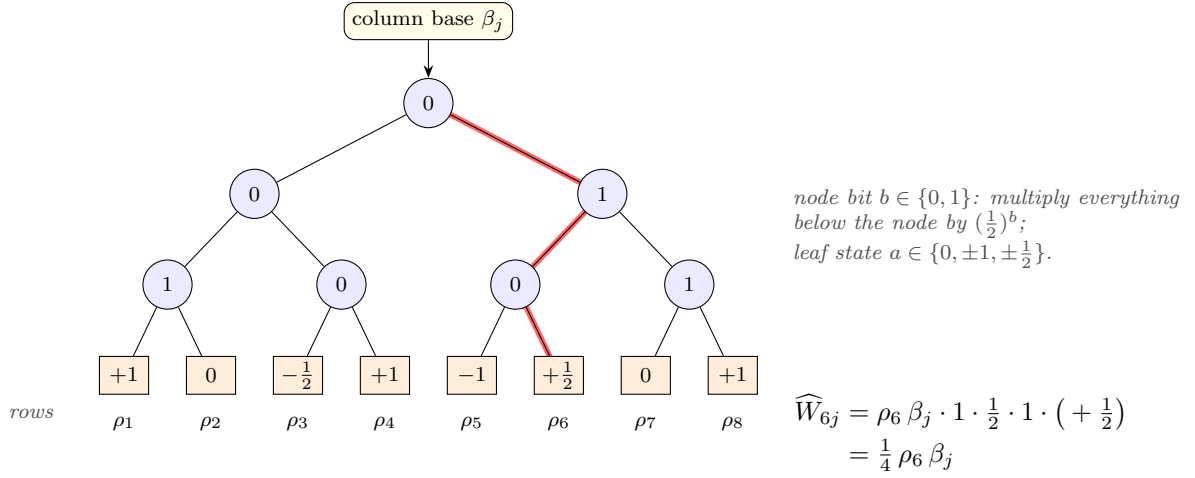

The example has a single shared node. A state tree applies the same idea at every level of a hierarchy over the entries of a matrix. \Cref{fig:anatomy} shows the tree we use as a running example. It describes one column of a weight matrix with eight rows, arranged as the leaves of a balanced binary tree. Every internal node holds one bit, which either leaves everything below it unchanged or halves it, and every leaf holds one of five states, $\{0,\pm1,\pm\frac12\}$: a sign and a small magnitude, possibly zero. The value of an entry is the product of the states on its path from the root, multiplied by a scale for its row and a base value for its column. For row~6, the path drawn in red gives $\frac14\rho_6\beta_j$.

In general, a state tree is a tree whose leaves are the entries of a matrix and whose nodes hold \emph{states}, each drawn from a small alphabet and each standing for a value; choosing the state of a node is one decision, and it is inherited by every entry below the node. The entry in row $i$ and column $j$ is
\begin{equation}\label{eq:value}
  \What_{ij}\;=\;\rho_i\,\beta_j\!\!\prod_{\text{nodes $v$ on the path to $(i,j)$}}\!\!\big(\text{value of the state at } v\big),
\end{equation}
where the row scales $\rho_i$ and column bases $\beta_j$ carry the overall scale of the matrix and may be stored in any format, for example FP16. Internal nodes stand for groups of entries (a block of a row, a pair of neighbouring rows, a channel, an attention head, the whole matrix), and a decision at a node is shared by its whole group. A tree can be as shallow as the single shared exponent of \cref{fig:example} or as deep as the running example. Because neighbours share decisions, the order of the rows matters: in the running example, rows with similar magnitudes are placed next to each other. A tree may group the entries of a row or of a column. The tree of \cref{fig:example} groups two entries of one row, so its shared shift acts once on a sum of inputs; the running example groups the entries of one column, so a shared shift acts once on an input before it fans out to several outputs. The two orientations cost the same bits, but they give different circuits, and, as \cref{sec:bm} will show, they couple different leaves when the states are optimized. A handful of kinds of states do most of the work:
\begin{itemize}[leftmargin=*]
\item a \emph{shift} has a power of two as its value and rescales everything below it, like the shared exponent of \cref{fig:example};
\item a \emph{gate} has the value zero and deletes everything below it, so nothing below a closed gate needs to be stored;
\item a \emph{transparent} state has the value one, so a node can always be present without changing anything;
\item \emph{signs and small magnitudes} complete the description at the leaves;
\item a \emph{lookup} is a state whose value depends on the states above it; for example, the entries below an index into a shared table read their values from the table.
\end{itemize}

\subsection{Combining trees}

A single tree describes every weight by one path. Three operations combine trees into richer descriptions, and each of them has a familiar meaning. The first is the \emph{sum}. Look again at the leaf of $w_1$ in \cref{fig:example}: its mantissa, $1.25=1+2^{-2}$, is already a sum of two powers of two, and each of the two terms became one wire into the adder. In general, a sum adds two trees entry by entry and expresses ``a coarse description plus a correction'': a first tree gives every weight its leading power of two, a second tree adds the next one where it is needed, and each additional term costs one addition. Sums are how a state tree buys precision. The second is the \emph{product}, which multiplies two trees as matrices. It expresses ``a product of two thin matrices'', as in low-rank factorization, or ``a matrix placed between two fixed transforms'', as in methods that rotate weights before quantizing them. The third is the \emph{reference}, which points to a tree defined elsewhere. It expresses ``look it up in a shared table'': the referenced tree is stored once and may be read by many entries, many matrices or many layers, as a codebook or a shared dictionary is. With references, the tree becomes a directed acyclic graph, which we call a \emph{state-tree graph}; we keep the shorter name when the distinction does not matter. Sums, products and references reappear in \cref{sec:express} as the dials that make the representation universal, and in \cref{sec:compress} as the edits behind familiar compression methods.

\subsection{Two prices: bits and additions}

The example of \cref{fig:example} already showed that a state tree has two costs, and both can be read from the tree itself. The \emph{storage cost} is the number of bits needed to write down the states: the logarithm of the alphabet size for every stored state, with shared parts counted once and nothing counted below a closed gate. In \cref{fig:example} it is 27 bits for two weights, against 32 in FP16. The column of \cref{fig:anatomy} stores seven node bits and eight five-state leaves, $7+8\log_25\approx25.6$ bits or about 3.2 bits per weight; for long columns the cost approaches $1+\log_25\approx3.3$ bits per weight, plus the row scales and column bases. Folding the root bit into the column base saves one more bit per column (\cref{prop:a-u2s1}), and a fractional cost such as $\log_25$ is reached by coding leaves jointly, for example three five-state leaves in 7 bits, since $5^3<2^7$. The \emph{circuit cost} is the number of additions needed to apply the matrix to an input when every state is a power of two, a short sum of powers of two, or zero. A power of two only shifts an input, a sign only negates it, and a zero removes it, so all that remains is adding up the shifted inputs: one addition for every term beyond the first in each output. In \cref{fig:example} this gives three additions. In the running example, every nonzero entry contributes a single shifted copy of its input to its output, and the two zero entries contribute nothing. \Cref{sec:circuits} develops this view for whole networks. The two prices need not agree, and the example shows why. Sharing the exponent of $w_1$ and $w_2$ saved five bits but not a single addition: a shift by three positions costs nothing in a circuit with fixed weights, whether it is applied once to the sum or separately to each term. The three additions came entirely from the nonzero digits at the leaves. In general, shared decisions save bits, while additions are saved only by having fewer nonzero terms, so the best tree for storage and the best tree for circuits can differ (\cref{sec:circuits}). Because the representation makes both prices explicit, an optimizer can trade one against the other.

\subsection{One tree, three readings}

The vocabulary is now complete, and it can be read in three ways (\cref{tab:readings}). In compression, the states are what is stored, and choosing them is the compression method. In a circuit, the states become wires and adders, and choosing them is the circuit design. In discrete optimization, the states are binary units, and choosing them is the search for the configuration of lowest energy. The rest of the paper follows these three readings in turn, after first establishing, in \cref{sec:express}, what state trees can and cannot express.

\begin{table}[t]
\centering
\small
\renewcommand{\arraystretch}{1.18}
\begin{tabularx}{\textwidth}{@{}L{2.75cm}YYY@{}}
\toprule
\textbf{Part of the tree} & \textbf{In compression} & \textbf{In a circuit} & \textbf{In a Boltzmann machine}\\
\midrule
Shift state & shared exponent or scale, as in \cref{fig:example} & a re-routing of wires & one binary unit\\
Zero state (gate) & pruning, from single weights to whole heads & a missing wire & a unit that switches off its
subtree\\
Leaf alphabet & quantization grid & a sign and a shift at an adder input & a one-hot group of units\\
Sum of trees & residual correction, extra precision & one more term, one more addition & one more term in each
weight\\
Product of trees & low rank, fixed rotations & two smaller circuits in sequence & units of two factors coupled
together\\
Shared subtree & codebook, dictionary, cross-layer sharing & stored once; computed once when its input is shared & units
shared by many weights\\
Choosing the states & the compression method & the circuit design & finding the ground state\\
\bottomrule
\end{tabularx}
\caption{One tree, three readings. Each part of a state tree has a meaning in compression
(\cref{sec:compress}), in circuits (\cref{sec:circuits}) and in discrete optimization
(\cref{sec:bm}); this is what lets one representation serve all three.}
\label{tab:readings}
\end{table}
\section{What state trees can express}\label{sec:express}
The example of \cref{fig:example} raises four natural questions. Does writing weights as state trees lose anything that today's formats offer? Can every matrix be described in this way, and at what price in bits? Can multiplications always be avoided without giving up precision? And what does it cost to share a decision? This section answers them in turn. We state each answer informally, as a property; the precise statements and their short proofs are collected in \cref{app:formal}. The properties refer to two variants of the model. In the \emph{core} model, the value of every state is fixed in advance. The general model also admits lookups, whose values depend on the states above them, as a mantissa depends on whether its exponent is subnormal. With lookups, any code that is decoded from the root down is a state tree, so \cref{prop:formats} is a statement about costs rather than about expressiveness. It is the only property that needs lookups (marked in \cref{tab:formats}); all others hold in the core model.

\subsection{Nothing is lost: today's formats are shallow state trees}

The shared exponent of \cref{fig:example} is not an exotic construction: it is what block floating point and the microscaling formats already do for blocks of weights, and a group scale in integer quantization plays the same role. Every format in common use turns out to be a state tree with only a few levels.

\begin{property}[Existing formats]\label{prop:formats}
Floating-point formats, integers with group scales and zero points, block floating point and microscaling formats (including shared microexponents), ternary weights, vector codebooks and $N{:}M$ sparsity are all shallow state trees, with at most five levels of decisions on any path. Each costs at most as many bits as the original format, and exactly
as many for the dense formats (\cref{tab:formats}; \cref{thm:a-formats}).
\end{property}

Adopting state trees is therefore a conservative step: whatever a format in use today can store, a state tree stores at no extra cost. Refining a tree never removes options either. A new node can always be given the transparent state, and an alphabet can always be enlarged, so a deeper tree contains every shallower one with the same leaves (\cref{prop:a-refine}). Most formats of \cref{tab:formats} occupy only the top and bottom of the tree; shared microexponents and HiFloat4 already use one or two levels in between, and these levels are where new structure can go.

\begin{table}[t]
\centering
\small
\renewcommand{\arraystretch}{1.15}
\begin{tabularx}{\textwidth}{@{}L{4.1cm}L{3.9cm}Y@{}}
\toprule
\textbf{Format} & \textbf{Shared decisions (nodes)} & \textbf{Private decisions (leaves)}\\
\midrule
FP32, FP16, BF16, FP8$^\dagger$ & --- & sign, exponent, mantissa (a chain of three)\\
Integers with group scale & scale per group & $b$-bit code\\
\quad with zero point$^\dagger$ & scale and zero point per group & $b$-bit code, read relative to the zero point\\
Block FP, microscaling (MX) & exponent per block & small float or integer\\
Shared microexponents (MX6, MX9), HiFloat4 & scale per block, then one or two levels of 1-bit shifts per sub-block & small float or integer\\
Ternary (BitNet b1.58) & one scale & $\{-1,0,+1\}$\\
Vector codebooks (e.g.\ QuIP\#)$^\dagger$ & index per block into a shared table & --- (entries read from the table)\\
$N{:}M$ sparsity$^\dagger$ & which $N$ of $M$ entries are nonzero & values of the $N$ kept entries\\
\bottomrule
\end{tabularx}
\caption{Formats in use are shallow state trees, at no more than their original cost (\cref{prop:formats}). $^\dagger$Uses lookups: subnormal mantissas, codes read relative to a zero point, entries read from a codebook, and entries kept by an $N{:}M$ mask; the other rows are core trees. Additive codebooks and rotations are built from these pieces with sums and products; trellis codes, whose overlapping windows share bits without nesting, fit only as a codebook over a whole block, with a table that is generated rather than stored. Deeper trees add levels of shared decisions between the two columns.}
\label{tab:formats}
\end{table}

\subsection{Nothing is lost in the worst case}

In \cref{fig:example}, sharing the exponent did not change which values the two weights could take; it only stopped the format from paying twice for the same decision. This is the general picture. What a tree can express is governed by the number of states it stores, and where those states sit decides how much the description costs.

\begin{property}[Completeness]\label{prop:complete}
A tree with $K$ possible values at each leaf represents every matrix whose entries come from a set of $K$ values, using $\log_2K$ bits per entry, and no encoding that can represent all such matrices uses fewer bits on every one of them. With finely spaced leaf values, state trees approximate every matrix to any accuracy, within a vanishing margin of the number of bits that any encoding needs in the worst case (\cref{thm:a-count,thm:a-density}).
\end{property}

A tree can always spend its states at the leaves and behave as an ordinary quantizer, so it never does worse than a flat format. Spending states higher up, where one state serves many weights, pays off when the weights have the corresponding structure: similar magnitudes within a group, as in \cref{fig:example}, runs of zeros, or repeated patterns. For matrices without any such structure, no representation can do better on all of them at once, and a state tree costs the same as a flat format. State trees do not create structure; they give it a place to go.

\subsection{No multiplications: shifts and additions suffice}

In \cref{fig:example}, every factor was a power of two, so every multiplication became a shift. The two mantissas helped: $1.25$ and $1.5$ each need only two power-of-two terms. A single tree whose states are powers of two can only produce values of the form $\pm2^k$ (\cref{fig:axes}a, middle). This is a logarithmic grid, and its worst-case relative error, one third, does not shrink as the tree grows. The remedy is the sum of \cref{sec:trees}: add a second tree for the next term, and a third if needed.

\begin{property}[Shifts and additions suffice]\label{prop:shift}
A sum of $t$ trees whose states are powers of two approximates every matrix to a relative error of $3^{-t}$ in every entry, within the range of exponents its states allow. It represents every matrix of integers of magnitude below $2^b$ exactly with $\lceil(b+1)/2\rceil$ terms per entry, and with about $b/3$ nonzero terms on average. Applying such a sum to an input needs shifts, additions, and only $m+n$ genuine multiplications, by the row scales and the column bases that the $t$ trees share (\cref{thm:a-shift}).
\end{property}

Each additional power-of-two term in a weight costs one addition and divides the worst-case error by three (\cref{fig:axes}a, bottom). Precision and circuit size thus become one and the same quantity, the number of terms, which \cref{sec:circuits} turns into circuits. The remaining $m+n$ multiplications can themselves be expanded into shifts and additions, or folded into neighbouring layers.

\subsection{The price of sharing}

Sharing a decision saves bits, but it constrains the weights that share it. In \cref{fig:example} the shared exponent required both weights to lie in the same octave. The halving bits of the running example relax this requirement in a precise way.

\begin{property}[What shared shifts allow]\label{prop:reach}
In a binary tree of shift bits, two nonzero entries whose closest common node lies $h$ levels above them can differ in magnitude by at most a factor of $2^h$. Every pattern of magnitudes that obeys this rule at every node, within the range that the depth of the tree allows, can be reached (\cref{prop:a-reach}).
\end{property}

Neighbours in the tree must therefore have similar magnitudes, which is why the order of the rows matters, and why a richer leaf alphabet or a second tree is used where neighbours differ a lot. The order matters for bits only; additions do not depend on it, since shifts are free in a fixed circuit. \Cref{prop:a-reach} also states what a good order must achieve: within every node, the exponents of the nonzero entries may spread by no more than the height of the node. Splitting the rows recursively into balanced halves of similar magnitude is a natural way to approach it. The rule is the same whether the shift bits sit on the edges of the tree, with leaves $\{0,\pm1\}$, or on its nodes, with leaves $\{0,\pm1,\pm\frac12\}$ as in the running example: the two encodings describe exactly the same columns, up to one factor per column (\cref{prop:a-u2s1}). Yet the second costs about 3.3 instead of 3.6 bits per weight, with leaves coded jointly in both. The saving comes from merging the shift on the edge into each leaf with the leaf's own alphabet: a zero needs no shift, so the six combinations of a shift bit and a three-state leaf collapse into five values. The same matrices can thus be described at different prices, and finding the cheaper description is part of choosing a tree. As a small proof of concept, we used the tree of the running example as a post-training quantizer for the MLP layers of LLaMA-3-8B, with one balanced tree per column over rows ordered by recursive bisection of their magnitude profiles, and states chosen by an exact dynamic program per column with GPTQ-style error feedback: at 3.33 bits per weight it reached a WikiText-2 perplexity of 7.78, against 8.03 for GPTQ at 3.25 bits,\footnote{Here only the MLP layers are quantized, and GPTQ uses groups of 128 weights. \Cref{tab:evidence} quantizes all linear layers of the transformer blocks with the strongest GPTQ configuration, which gives 8.09.} and the edge-bit encoding produced identical weights at 3.60 bits, as the equivalence of the two encodings predicts.

\subsection{Beyond states: rank and sharing}

\begin{figure}[t]
\centering
\begin{subfigure}[b]{0.99\textwidth}
\centering
\begin{tikzpicture}[x=4.9cm,y=1cm]
\draw[-{Stealth}] (-1.08,0) -- (1.1,0) node[right,font=\scriptsize] {value};
\foreach \v in {-1,-0.714,-0.429,-0.143,0.143,0.429,0.714,1} {\fill[blue!70] (\v,0) circle (2.1pt);}
\node[anchor=east,font=\scriptsize,align=right] at (-1.12,0) {flat 3-bit grid\\8 states};
\foreach \t in {-1,0,1} {\draw (\t,-0.07) -- (\t,0.07); \node[below,font=\tiny] at (\t,-0.07) {$\t$};}
\begin{scope}[yshift=-1.05cm]
\draw[-{Stealth}] (-1.08,0) -- (1.1,0);
\foreach \v in {-1,-0.5,-0.25,-0.125,0,0.125,0.25,0.5,1} {\fill[red!70] (\v,0) circle (2.1pt);}
\node[anchor=east,font=\scriptsize,align=right] at (-1.12,0) {shift tree, depth 3\\values $\pm2^{-k}$ and $0$};
\foreach \t in {-1,0,1} {\draw (\t,-0.07) -- (\t,0.07);}
\end{scope}
\begin{scope}[yshift=-2.1cm]
\draw[-{Stealth}] (-1.08,0) -- (1.1,0);
\foreach \k in {-8,...,8} {\fill[green!50!black] (\k/8,0) circle (2.1pt);}
\node[anchor=east,font=\scriptsize,align=right] at (-1.12,0) {sum of 2 shift trees\\17 levels, $\le2$ terms};
\foreach \t in {-1,0,1} {\draw (\t,-0.07) -- (\t,0.07);}
\end{scope}
\end{tikzpicture}
\caption{States: the values one entry can take.}
\end{subfigure}

\vspace{4mm}
\begin{subfigure}[b]{0.46\textwidth}
\centering
\begin{tikzpicture}
\node[opnode] (s) at (0,0) {$+$};
\foreach \p/\k/\x in {p1/1/-2.3,p2/2/0,p3/r/2.3} {
  \node[opnode] (\p) at (\x,-1.15) {$\times$};
  \draw (s) -- (\p);
  \node[leaf] (u\k) at (\x-0.45,-2.2) {$u_{\k}$};
  \node[leaf] (v\k) at (\x+0.45,-2.2) {$v_{\k}^{\!\top}$};
  \draw (\p) -- (u\k); \draw (\p) -- (v\k);}
\node[font=\small] at (1.15,-1.15) {$\cdots$};
\end{tikzpicture}
\caption{Rank: $\What=\sum_{k\le r}u_kv_k^\top$.}
\end{subfigure}\hfill
\begin{subfigure}[b]{0.5\textwidth}
\centering
\begin{tikzpicture}
\foreach \w/\x/\t in {w1/0/{$W_1$ (layer 3)},w2/2.3/{$W_2$ (layer 3)},w3/4.6/{$W_3$ (layer 17)}} {
  \node[opnode] (\w) at (\x,0) {$\times$};
  \node[font=\scriptsize, above=1mm of \w] {\t};}
\foreach \c/\x in {c1/-0.75,c2/1.55,c3/3.85} {\node[leaf] (\c) at (\x,-0.95) {};}
\node[font=\scriptsize] at (c1) {$C_1$}; \node[font=\scriptsize] at (c2) {$C_2$}; \node[font=\scriptsize] at (c3) {$C_3$};
\node[shared] (D) at (2.3,-2.35) {shared subtree (stored once)};
\draw (w1) -- (c1); \draw (w2) -- (c2); \draw (w3) -- (c3);
\draw[-{Stealth},thick,purple] (w1.south east) -- (D.north west);
\draw[-{Stealth},thick,purple] (w2.south) -- (D.north);
\draw[-{Stealth},thick,purple] (w3.south) -- (D.north east);
\end{tikzpicture}
\caption{Sharing: several matrices refer to one stored subtree.}
\end{subfigure}
\caption{Three dials along which state trees grow. (a)~Adding states: a flat grid; a shift tree, whose values form a logarithmic grid; and a sum of two shift trees, which reaches the uniform 17-level grid of all multiples of $\frac18$ in $[-1,1]$, since each of them is a sum of at most two signed powers of two with exponents down to $-3$ (\cref{prop:shift}). (b)~Adding rank. (c)~Sharing a subtree among several matrices, possibly in different layers. With real-valued factors, rank or sharing alone reaches every matrix.}
\label{fig:axes}
\end{figure}
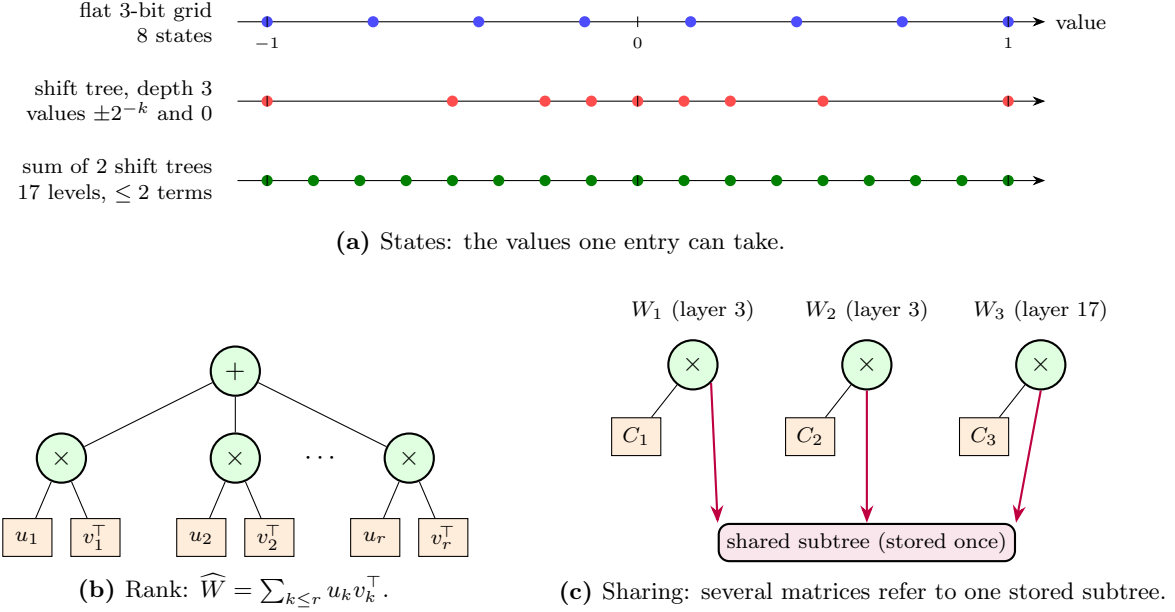

The number of states is one dial along which a state tree grows. The other two operations of \cref{sec:trees} provide two more (\cref{fig:axes}b,c). \emph{Rank.} A sum of $r$ products of a column and a row describes exactly the matrices of rank at most $r$, so full rank describes every matrix. When the error is measured on the outputs of the layer, the best low-rank tree has a closed form, a singular value decomposition in a reweighted space, a classical result of reduced-rank regression \cite{izenman1975rrr} that SVD-LLM \cite{wang2025svdllm} recovers (\cref{thm:a-rank}). \emph{Sharing.} Several matrices with the same input size, in one layer or in different layers, can refer to one shared subtree that is stored once (\cref{fig:axes}c). The more the shared part contains, the more the matrices have in common, and a large enough shared part describes them exactly (\cref{prop:a-share}). With real-valued factors, each of these two dials on its own reaches every matrix. When the factors are themselves written as state trees, the three dials combine, and a single description can trade states, rank and sharing against one another.

\paragraph{Summary.}
State trees lose nothing that today's formats offer, they lose nothing in the worst case, they reach any precision with shifts and additions, and they make the price of sharing explicit. What they offer in return is a single object on which the many methods of model compression can be expressed and compared, which is the subject of section 5.
\section{Building a tree, step by step}\label{sec:example}

The previous sections described state trees in general. In this section, we illustrate how to build one in full, on a small matrix, so that every decision can be followed. The tree we build is the one behind our lossless format (\cref{sec:directions}): it stores BF16 weights exactly, in fewer bits, and a GPU decodes it directly into the registers of its tensor cores. Every choice below is made with that decoder in mind.

\begin{figure}[t]
\centering
\begin{subfigure}[b]{0.42\textwidth}
\centering
\begin{tikzpicture}[x=1cm,y=1cm]
\tikzset{
  wcell/.style={draw=black!45, minimum width=1.2cm, minimum height=0.56cm, inner sep=0pt,
    font=\scriptsize},
  o3/.style={fill=teal!46}, o4/.style={fill=teal!32}, o5/.style={fill=teal!20},
  o6/.style={fill=teal!10}, o7/.style={fill=teal!3},
}
\foreach \i/\j/\v/\o in {
  1/1/{$\phantom{-}0.0391$}/o5, 1/2/{$-0.0234$}/o6, 1/3/{$\phantom{-}0.0352$}/o5, 1/4/{$-0.0137$}/o7,
  2/1/{$-0.0703$}/o4, 2/2/{$\phantom{-}0.0215$}/o6, 2/3/{$\phantom{-}0.1055$}/o4, 2/4/{$-0.0352$}/o5,
  3/1/{$\phantom{-}0.0195$}/o6, 3/2/{$-0.0088$}/o7, 3/3/{$-0.0469$}/o5, 3/4/{$\phantom{-}0.0283$}/o6,
  4/1/{$\phantom{-}0.1875$}/o3, 4/2/{$-0.0156$}/o6, 4/3/{$\phantom{-}0.0410$}/o5, 4/4/{$-0.0625$}/o4} {
  \node[wcell, \o] at ({1.2*\j}, {-0.56*\i}) {\v};}
\foreach \i in {1,...,4} {
  \draw[blue!60!black, rounded corners=2pt, thick]
    (0.57, {-0.56*\i+0.25}) rectangle (5.43, {-0.56*\i-0.25});
  \node[font=\scriptsize, blue!60!black, anchor=west] at (5.48, {-0.56*\i}) {$g_{\i}$};}
\node[font=\scriptsize, text=black!70] at (3.0, -0.02) {inputs $j\;\longrightarrow$};
\node[font=\scriptsize, text=black!70, rotate=90] at (0.3, -1.4) {outputs $i$};
\foreach \o/\l [count=\c from 0] in {o3/{-3}, o4/{-4}, o5/{-5}, o6/{-6}, o7/{-7}} {
  \node[draw=black!45, \o, minimum width=0.36cm, minimum height=0.28cm, inner sep=0pt]
    at ({1.0+1.0*\c}, -2.95) {};
  \node[font=\tiny, anchor=west] at ({1.2+1.0*\c}, -2.95) {$2^{\l}$};}
\node[font=\tiny, text=black!70, anchor=east] at (0.8, -2.95) {octave};
\end{tikzpicture}
\caption{A $4\times4$ matrix in BF16 (values rounded); shading shows each weight's octave.}
\end{subfigure}\hfill
\begin{subfigure}[b]{0.56\textwidth}
\centering
\scriptsize
\renewcommand{\arraystretch}{1.25}
\begin{tabular}{@{}l@{\ \ }c@{\ \ }c@{\ \ }c@{\quad}c@{\ \ }c@{}}
\toprule
& \multicolumn{3}{c}{\textbf{BF16 fields}} & \multicolumn{2}{c}{\textbf{in the tree}}\\
\cmidrule(r){2-4}\cmidrule(l){5-6}
weight & $s$ & $e$ & $m$ & $D$ & leaf\\
\midrule
$w_{11}=\phantom{-}0.0391$ & 0 & 122 ($2^{-5}$) & \texttt{0100000} & 0 & $+1.25$\\
$w_{12}=-0.0234$ & 1 & 121 ($2^{-6}$) & \texttt{1000000} & 1 & $-1.5$\\
$w_{13}=\phantom{-}0.0352$ & 0 & 122 ($2^{-5}$) & \texttt{0010000} & 0 & $+1.125$\\
$w_{14}=-0.0137$ & 1 & 120 ($2^{-7}$) & \texttt{1100000} & 2 & $-1.75$\\
\midrule
\multicolumn{6}{@{}l@{}}{group $g_1$: shared exponent $E=\max e=122$ ($2^{-5}$); state $D=E-e$}\\
\bottomrule
\end{tabular}
\vspace{2mm}
\caption{Row 1: each weight's three BF16 decisions, and what the tree keeps of them.}
\end{subfigure}

\vspace{5mm}
\begin{subfigure}[b]{\textwidth}
\centering
\begin{tikzpicture}[x=1cm,y=1cm]
\tikzset{
  gnode/.style={draw, rounded corners=2pt, fill=blue!12, font=\scriptsize, inner sep=2.5pt,
    minimum height=0.5cm},
  snode/.style={circle, draw, fill=orange!10, font=\scriptsize, minimum size=0.46cm, inner sep=0pt},
  lnode/.style={draw, fill=orange!18, font=\fontsize{5}{6}\selectfont, minimum width=0.76cm,
    minimum height=0.42cm, inner sep=1pt},
}
\def\dx{0.985}
\foreach \D/\L [count=\k from 0] in {
  0/{$+1.25$}, 1/{$-1.5$}, 0/{$+1.125$}, 2/{$-1.75$},
  0/{$-1.125$}, 2/{$+1.375$}, 0/{$+1.6875$}, 1/{$-1.125$},
  1/{$+1.25$}, 2/{$-1.125$}, 0/{$-1.5$}, 1/{$+1.8125$},
  0/{$+1.5$}, 3/{$-1.0$}, 2/{$+1.3125$}, 1/{$-1.0$}} {
  \node[snode] (s\k) at ({\dx*\k}, -2.5) {\D};
  \node[lnode] (l\k) at ({\dx*\k}, -3.55) {\L};
  \draw (s\k) -- (l\k);}
\foreach \g/\E [count=\c from 0] in {1/{-5}, 2/{-4}, 3/{-5}, 4/{-3}} {
  \node[gnode] (g\g) at ({\dx*(4*\c+1.5)}, -1.25) {$2^{\E}$};
  \pgfmathtruncatemacro{\a}{4*\c}
  \foreach \t in {0,...,3} {
    \pgfmathtruncatemacro{\k}{\a+\t}
    \draw (g\g) -- (s\k);}
  \node[font=\scriptsize, text=black!70] at ({\dx*(4*\c+1.5)}, -4.1) {row \g\ $=$ group $g_{\g}$};}
\node[circle, draw, dashed, fill=white, font=\scriptsize, minimum size=0.55cm, inner sep=0pt] (r)
  at ({\dx*7.5}, 0) {$W$};
\foreach \g in {1,...,4} {\draw[dashed, black!60] (r) -- (g\g);}
\begin{scope}[on background layer]
\draw[line width=4.5pt, red!30, line cap=round, line join=round]
  (r.center) -- (g1.center) -- (s3.center) -- (l3.center);
\end{scope}
\node[font=\scriptsize, red!60!black, anchor=west] at (-0.38, 0.3)
  {path of $w_{14}$: $\;2^{-5}\cdot2^{-2}\cdot(-1.75)=-0.013671875$};
\begin{scope}[yshift=-4.75cm, font=\scriptsize]
\node[circle, draw, dashed, minimum size=0.3cm, inner sep=0pt] at (0.6,0) {};
\node[anchor=west] at (0.85,0) {shape of $W$: fixed, not stored};
\node[draw, rounded corners=1.5pt, fill=blue!12, minimum width=0.4cm, minimum height=0.3cm] at (7.9,0) {};
\node[anchor=west] at (8.2,0) {shared exponent $E$: 8 bits per group};
\node[circle, draw, fill=orange!10, minimum size=0.3cm, inner sep=0pt] at (0.6,-0.45) {};
\node[anchor=west] at (0.85,-0.45) {state $D$, a shift by $2^{-D}$: 4 bits per weight};
\node[draw, fill=orange!18, minimum width=0.4cm, minimum height=0.3cm] at (7.9,-0.45) {};
\node[anchor=west] at (8.2,-0.45) {leaf: sign and mantissa, 8 bits per weight};
\end{scope}
\end{tikzpicture}
\caption{The state tree of the whole matrix. The path of $w_{14}$ (red) multiplies the shared
exponent, the shift of its state and its leaf, and returns the BF16 value exactly.}
\end{subfigure}
\caption{Building the tree of our lossless format. (a)~Weights of a trained layer lie within a few octaves of one another, but rarely share an exponent exactly. (b)~Each group keeps its largest exponent once; every weight keeps only its distance $D$ below it, and its sign and mantissa. (c)~The resulting tree has a fixed shape: one node per group, one state and one leaf per weight. Blue: shared decisions; orange: private decisions. The example uses groups of 4 weights; the format uses groups of 64.}
\label{fig:build}
\end{figure}

\paragraph{Step 1: every weight is a path.}
\Cref{fig:build}a shows a $4\times4$ matrix in BF16. As in \cref{sec:trees}, each weight is read as three decisions: a sign $s$, an 8-bit exponent field $e$ and a 7-bit mantissa $m$, with $w=(-1)^s\cdot2^{\,e-127}\cdot(1+m/128)$. The shading shows the octave $2^{e-127}$ of every weight. Within a row the weights lie within a few octaves of one another, as trained weights usually do, but their exponents rarely coincide, so the single shared exponent of \cref{fig:example} cannot be used as it stands.
\paragraph{Step 2: choose the groups.}
The tree groups the four weights of every row in their natural order; in the format a group holds 64 consecutive weights of a row. Groups run along the input dimension for two reasons. Their weights are multiplied by consecutive inputs and summed into the same output, so in a circuit the shared factor of a group could be applied once to its partial sum, as in \cref{fig:example}c. And they lie next to one another in memory, in the order in which a tensor core reads them. Because the order is the natural one, no permutation has to be stored or applied. \paragraph{Step 3: lift the shared decision.} Every group receives a node that holds its largest exponent field $E$, stored once in 8 bits. Below it, every weight keeps only its distance to that exponent, the state $D=E-e$, which stands for a shift by $2^{-D}$ and takes 4 bits, and a leaf that holds its sign and mantissa in 8 bits (\cref{fig:build}b). Along every path the product is unchanged,
\[
  w\;=\;2^{\,E-127}\cdot2^{-D}\cdot(-1)^s\Big(1+\frac{m}{128}\Big),
\]
so the tree reproduces every weight exactly. In row~1, $E$ encodes $2^{-5}$ and the four states are 0, 1, 0 and~2; the red path in \cref{fig:build}c decodes $w_{14}$. Unlike the shared exponent of \cref{sec:trees}, this node does not require the exponents of a group to be equal: the states absorb their differences, which is the second kind of shared decision anticipated there. \paragraph{Step 4: set aside the rare outliers.} A 4-bit state distinguishes 16 values. The states 0 to 14 are distances, so the weights of a group may spread over 15 octaves. A weight that lies further below the largest of its group, an exact zero for instance, takes the state 15 as an escape: its leaf is left empty, and its 16 bits are kept exactly in a short side list. No weight of the example needs an escape; in the models we tested, about one weight in ten thousand does (\cref{app:exp}).
\paragraph{Step 5: count the bits.}
A group stores $8+4\times(4+8)=56$ bits for its four weights, against 64 in BF16. As with the shared exponent of \cref{sec:trees}, the saving grows with the group: with 64 weights a group needs $8/64+12=12.125$ bits per weight, and with the escapes, whole models take 12.13 bits per weight, 24\% less than BF16, with every weight bit-exact.

\begin{figure}[t]
\centering
\begin{tikzpicture}[x=1cm,y=1cm]
\tikzset{
  fld/.style={draw, minimum height=0.55cm, inner sep=0pt, font=\scriptsize\ttfamily},
  lab/.style={font=\scriptsize},
  brc/.style={decorate, decoration={brace, amplitude=4pt}, black!60},
}
\node[fld, fill=blue!12, minimum width=2cm] (E) at (1,0) {01111010};
\foreach \d/\x [count=\c] in {0000/2.5, 0001/3.5, 0000/4.5, 0010/5.5} {
  \node[fld, fill=orange!10, minimum width=1cm] (D\c) at (\x,0) {\d};}
\foreach \l/\x [count=\c] in {{0\,0100000}/7, {1\,1000000}/9, {0\,0010000}/11} {
  \node[fld, fill=orange!18, minimum width=2cm] (L\c) at (\x,0) {\l};}
\node[fld, fill=orange!18, minimum width=2cm] (L4) at (13,0)
  {\textcolor{red!70!black}{1}\,\textcolor{violet}{1100000}};
\node[lab] at (1,0.45) {$E$};
\foreach \c in {1,...,4} {\node[lab] at ($(D\c)+(0,0.45)$) {$D_{\c}$};
                          \node[lab] at ($(L\c)+(0,0.45)$) {$L_{\c}$};}
\draw[brc] (0.05,0.72) -- (1.95,0.72) node[midway, above=4pt, lab, align=center]
  {byte 0\\[-1pt]\textcolor{black!60}{shared exponent}};
\draw[brc] (2.05,0.72) -- (5.95,0.72) node[midway, above=4pt, lab, align=center]
  {bytes 1--2\\[-1pt]\textcolor{black!60}{states, 4 bits each}};
\draw[brc] (6.05,0.72) -- (13.95,0.72) node[midway, above=4pt, lab, align=center]
  {bytes 3--6\\[-1pt]\textcolor{black!60}{leaves: sign and mantissa, 8 bits each}};
\node[circle, draw, fill=green!12, minimum size=0.6cm, inner sep=0pt, font=\small] (sub) at (3.5,-1.45) {$-$};
\draw[-{Stealth}] (1,-0.29) |- (sub.west);
\draw[-{Stealth}] (5.5,-0.29) |- (sub.east);
\node[lab, text=black!70, anchor=west] at (3.85,-1.95) {$122-2=120$};
\node[fld, fill=red!8, minimum width=0.5cm] (rs) at (10.0,-2.75) {\textcolor{red!70!black}{1}};
\node[fld, fill=green!12, minimum width=2cm] (re) at (11.25,-2.75) {01111000};
\node[fld, fill=violet!8, minimum width=1.75cm] (rm) at (13.125,-2.75) {\textcolor{violet}{1100000}};
\node[lab, anchor=east] at (9.6,-2.62) {BF16 pattern of $w_{14}$};
\node[lab, anchor=east] at (9.6,-3.02) {$=\texttt{0xBC60}=-2^{-7}\cdot1.75=-0.013671875$};
\draw[-{Stealth}] (sub.south) -- (3.5,-3.6) -| (re.south);
\draw[-{Stealth}, dashed, red!70!black] (12.15,-0.29) -- (12.15,-1.05) -| (rs.north);
\draw[-{Stealth}, dashed, violet] (13.35,-0.29) -- (13.35,-2.475);
\node[lab, text=black!70, align=center] at (11.6,-1.75) {copied\\[-1pt]from $L_4$};
\node[lab, anchor=west, text=black!75, align=left] at (0,-4.35)
  {\texttt{bf16 = (s\,<<\,15) | ((E\,-\,D)\,<<\,7) | m}, written directly into the tensor-core registers;\\[2pt]
   $D_j$ lies at byte $1+\lfloor (j-1)/2\rfloor$ and $L_j$ at byte $2+j$: fixed offsets, no pointers.};
\end{tikzpicture}
\caption{The stored record of group $g_1$ and the decoding of $w_{14}$. Every field has a fixed width at a fixed offset, so the address of any weight's state and leaf follows from its index alone. Decoding subtracts the state from the shared exponent and copies the sign and mantissa from the leaf. In the format, a block of 16 rows and 64 columns, the shape the kernel feeds to the tensor cores, is one record of $16\times97=1{,}552$ bytes.}
\label{fig:decode}
\end{figure}
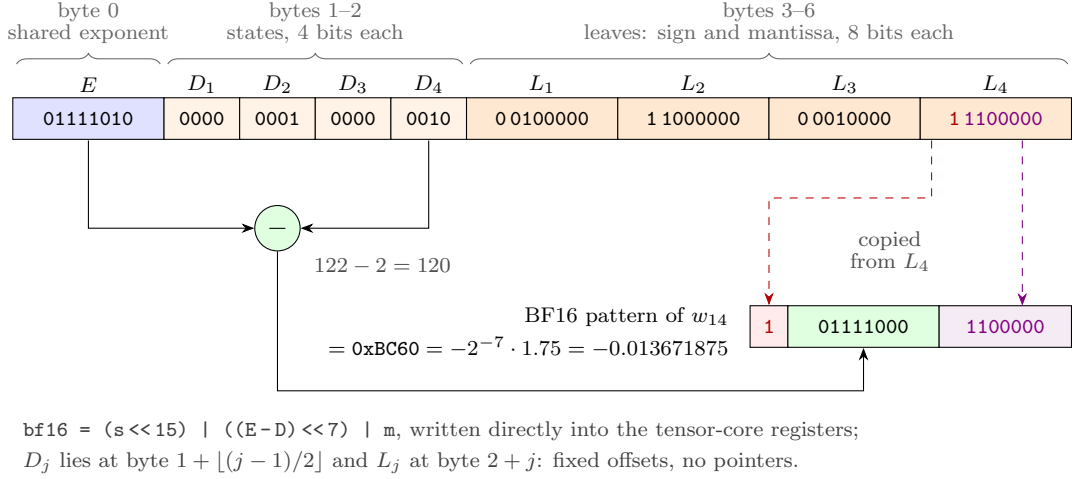

\paragraph{Step 6: decode without pointers.}
\Cref{fig:decode} shows how the first group is stored. Every field has a fixed width at a fixed offset: the shared exponent in the first byte, the four states in the next two and the four leaves in the last four. A thread that needs weight $j$ of a group computes the addresses of $D_j$ and $L_j$ from $j$ alone; it never reads a pointer and never branches, except at the rare escape. Rebuilding the BF16 pattern takes a subtraction and two copies: the exponent field is $E-D_j$, and the sign and mantissa come from the leaf. The result is written directly into the registers from which the tensor cores read, so the weights never travel through memory in BF16. This is what lets the kernel of \cref{sec:directions} run at the speed of memory.

\paragraph{Why a tree, and not something smaller.}
The tree we have built is not the shortest possible description of the matrix. Groups whose subtrees coincide could be stored once, turning the tree into a graph; prefixes shared by different groups could be merged, as in a trie; weights could be grouped by magnitude rather than by position; and the states could be entropy-coded, since small distances are far more frequent than large ones. Each of these would lower the bits further, but each needs pointers, indices or codes of variable length that a decoder must follow before it can locate a weight, and that breaks the fixed addressing that lets a tensor core read its weights at full speed. We therefore keep the fixed-shape tree and leave these richer representations to future work. We do not claim that this tree, or trees in general, are the optimal representation of a weight matrix. We claim that the tree improves on the representations in use in several respects at once: it stores exactly the same values in fewer bits, it decodes at the speed of memory, and it makes explicit the structure that the following sections edit, wire and optimize.

\paragraph{From exact to compressed.}
The same tree also shows what compression does. Keeping only the leading three bits of every mantissa shrinks the leaf alphabet from 256 to 16 values and the leaf from 8 to 4 bits. The tree keeps its shape, but now describes an approximation of the matrix rather than the matrix itself. Every compression method can be read as an edit of this kind, which is where the next section begins.

\section{Compression is editing one tree}\label{sec:compress}

Model compression has grown into several families (pruning, quantization, low-rank factorization, codebooks and weight sharing), each with its own format, its own solver and often its own kernel. Combining them is hard precisely because their formats differ. In a state tree they become something simpler: different ways of editing one object.

\subsection{Every method is an edit}

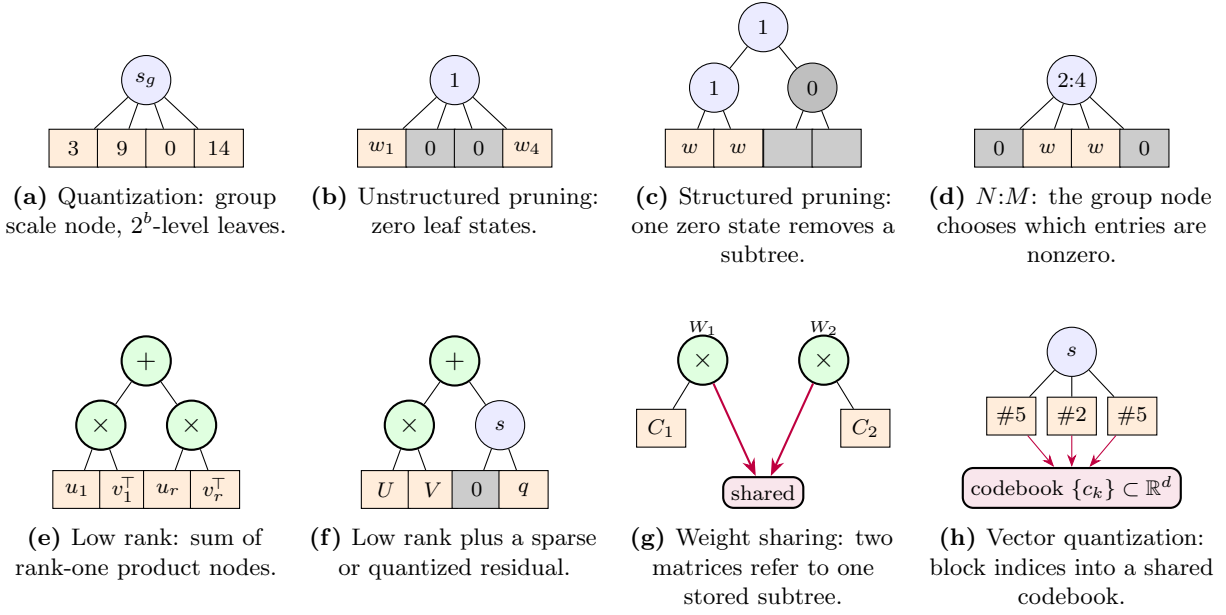
\begin{figure}[t]
\centering
\begin{subfigure}[t]{0.235\textwidth}\centering
\begin{tikzpicture}[level distance=9mm, level 1/.style={sibling distance=6.4mm}]
\node[inode] {$s_g$}
  child {node[leaf] {$3$}} child {node[leaf] {$9$}} child {node[leaf] {$0$}} child {node[leaf] {$14$}};
\end{tikzpicture}
\caption{Quantization: group scale node, $2^b$-level leaves.}
\end{subfigure}\hfill
\begin{subfigure}[t]{0.235\textwidth}\centering
\begin{tikzpicture}[level distance=9mm, level 1/.style={sibling distance=6.4mm}]
\node[inode] {$1$}
  child {node[leaf] {$w_1$}} child {node[zleaf] {$0$}} child {node[zleaf] {$0$}} child {node[leaf] {$w_4$}};
\end{tikzpicture}
\caption{Unstructured pruning: zero leaf states.}
\end{subfigure}\hfill
\begin{subfigure}[t]{0.235\textwidth}\centering
\begin{tikzpicture}[level distance=8mm, level 1/.style={sibling distance=13mm}, level 2/.style={sibling distance=6.4mm}]
\node[inode] {$1$}
  child {node[inode] {$1$} child {node[leaf] {$w$}} child {node[leaf] {$w$}}}
  child {node[znode] {$0$} child {node[zleaf] {}} child {node[zleaf] {}}};
\end{tikzpicture}
\caption{Structured pruning: one zero state removes a subtree.}
\end{subfigure}\hfill
\begin{subfigure}[t]{0.235\textwidth}\centering
\begin{tikzpicture}[level distance=9mm, level 1/.style={sibling distance=6.4mm}]
\node[inode] {$2{:}4$}
  child {node[zleaf] {$0$}} child {node[leaf] {$w$}} child {node[leaf] {$w$}} child {node[zleaf] {$0$}};
\end{tikzpicture}
\caption{$N{:}M$: the group node chooses which entries are nonzero.}
\end{subfigure}

\vspace{5mm}
\begin{subfigure}[t]{0.235\textwidth}\centering
\begin{tikzpicture}[level distance=8.5mm, level 1/.style={sibling distance=12mm}, level 2/.style={sibling distance=6.2mm}]
\node[opnode] {$+$}
  child {node[opnode] {$\times$} child {node[leaf] {$u_1$}} child {node[leaf] {$v_1^{\!\top}$}}}
  child {node[opnode] {$\times$} child {node[leaf] {$u_r$}} child {node[leaf] {$v_r^{\!\top}$}}};
\end{tikzpicture}
\caption{Low rank: sum of rank-one product nodes.}
\end{subfigure}\hfill
\begin{subfigure}[t]{0.235\textwidth}\centering
\begin{tikzpicture}[level distance=8.5mm, level 1/.style={sibling distance=12mm}, level 2/.style={sibling distance=6.2mm}]
\node[opnode] {$+$}
  child {node[opnode] {$\times$} child {node[leaf] {$U$}} child {node[leaf] {$V$}}}
  child {node[inode] {$s$} child {node[zleaf] {$0$}} child {node[leaf] {$q$}}};
\end{tikzpicture}
\caption{Low rank plus a sparse or quantized residual.}
\end{subfigure}\hfill
\begin{subfigure}[t]{0.235\textwidth}\centering
\begin{tikzpicture}
\node[opnode] (a) at (0,0) {$\times$};
\node[opnode] (b) at (1.6,0) {$\times$};
\node[leaf] (ca) at (-0.55,-0.9) {$C_1$};
\node[leaf] (cb) at (2.15,-0.9) {$C_2$};
\node[shared] (d) at (0.8,-1.75) {shared};
\draw (a) -- (ca); \draw (b) -- (cb);
\draw[-{Stealth},thick,purple] (a) -- (d); \draw[-{Stealth},thick,purple] (b) -- (d);
\node[font=\tiny] at (0,0.42) {$W_1$}; \node[font=\tiny] at (1.6,0.42) {$W_2$};
\end{tikzpicture}
\caption{Weight sharing: two matrices refer to one stored subtree.}
\end{subfigure}\hfill
\begin{subfigure}[t]{0.235\textwidth}\centering
\begin{tikzpicture}
\node[inode] (r) at (0.8,0) {$s$};
\node[leaf] (b1) at (0,-0.85) {$\#5$};
\node[leaf] (b2) at (0.8,-0.85) {$\#2$};
\node[leaf] (b3) at (1.6,-0.85) {$\#5$};
\draw (r) -- (b1); \draw (r) -- (b2); \draw (r) -- (b3);
\node[shared] (cb) at (0.8,-1.8) {codebook $\{c_k\}\subset\R^d$};
\draw[-{Stealth},purple] (b1) -- (cb); \draw[-{Stealth},purple] (b2) -- (cb); \draw[-{Stealth},purple] (b3) -- (cb);
\end{tikzpicture}
\caption{Vector quantization: block indices into a shared codebook.}
\end{subfigure}
\caption{Compression methods as edits of a state-tree graph. Grey: zero states; green: sum ($+$)
and product ($\times$) nodes; purple: references to shared subtrees.}
\label{fig:gallery}
\end{figure}

The two weights of \cref{fig:example} already show most of the edits. Replacing the leaf of $w_2$ by the state $0$ prunes $w_2$; a zero state at the shared node removes both weights at once, which is structured pruning in miniature. Shortening the mantissas is quantization: $1.25$ and $1.5$ use only two of their ten mantissa bits, so for these two weights a two-bit alphabet loses nothing. The shared exponent itself is an inserted node, the same edit that produces group scales and block formats. Writing $1.25$ as $1+2^{-2}$ splits a leaf into a sum, which is how residual corrections and additive codebooks work. The remaining edits act on whole matrices: factoring a tree into a product gives low rank and rotations, a reference to a shared part gives codebooks and weight sharing across layers, and a selector that chooses among several subtrees gives mixed precision and the choice of a method for each matrix. \Cref{tab:edits} lists these seven edits, and \cref{fig:gallery} draws the most common ones. Common compression methods, including all those cited in the table, are combinations of them, together with a rule for choosing the new states.

\begin{table}[t]
\centering
\small
\renewcommand{\arraystretch}{1.15}
\begin{tabularx}{\textwidth}{@{}L{3.4cm}L{4.5cm}Y@{}}
\toprule
\textbf{Edit} & \textbf{What changes in the tree} & \textbf{Familiar name}\\
\midrule
Zero a state & a leaf or node state takes the value $0$ & unstructured, $N{:}M$ and structured pruning
\cite{lecun1990obd,hassibi1993obs,frantar2023sparsegpt,sun2024wanda,meng2024alps,shopkhoev2025replaceme}\\
Shrink an alphabet & fewer states at the leaves or nodes & quantization \cite{frantar2023gptq,lin2024awq,kim2024squeezellm},
ternarization \cite{ma2024era}\\
Insert a shared node & a new group with a scale, shift or sign & group scales, shared exponents
\cite{rouhani2023mx}\\
Split into a sum & $W\to W_1+W_2$ & low rank plus residual, additive codebooks \cite{egiazarian2024aqlm}\\
Factor into a product & $W\to UV$ or $W\to R_1W'R_2$ & low rank \cite{wang2025svdllm,thoma2026lems,ali2026rocket,makhov2026compot}, rotations
\cite{tseng2024quipsharp}, butterfly \cite{dao2019butterfly}\\
Reference a shared part & a subtree is stored once and used many times & codebooks
\cite{vanbaalen2024gptvq,tseng2024quipsharp,tseng2024qtip}, sparse dictionaries \cite{makhov2025cospadi}, cross-layer sharing \cite{lan2020albert,wang2025basissharing,zhussip2026masa,mohammad2026geopair}\\
Choose among subtrees & a selector state picks one encoding & mixed precision, method selection\\
\bottomrule
\end{tabularx}
\caption{Seven edits of a state-tree graph. Common compression methods, including those cited here, are
combinations of these edits, together with a rule for choosing the new states.}
\label{tab:edits}
\end{table}

\subsection{Methods differ only in how they choose states}

Seen this way, methods within one family share a representation and differ only in how they choose the states. Magnitude pruning, Wanda \cite{sun2024wanda}, SparseGPT \cite{frantar2023sparsegpt} and ALPS \cite{meng2024alps} all produce zero states at the leaves; they differ in their criterion and their solver. Round-to-nearest, GPTQ \cite{frantar2023gptq} and AWQ \cite{lin2024awq} all choose leaf states under group-scale nodes; again, only the rule for choosing differs. Two consequences follow. Methods of the same family can be compared on equal terms, since they produce the same kind of object. And methods from different families can be combined without inventing a new format: a quantized low-rank factor, or a pruned tree that reads from a shared codebook, is simply a deeper tree.

\subsection{Pricing an edit}

To choose among edits, each one needs a price in accuracy. Because every edit changes the same kind of object, a single error measure can price all of them. The standard choice measures the error on the outputs of the layer: with calibration inputs $x_1,\dots,x_N\in\R^n$, taken as column vectors, their second moment $H=\frac1N\sum_\nu x_\nu x_\nu^\top$ and the error $E=W-\What$,
\begin{equation}\label{eq:fH}
  f_H(\What)\;=\;\tr\big(EHE^\top\big)\;=\;\frac1N\sum_{\nu=1}^N\big\|Ex_\nu\big\|^2 .
\end{equation}
Correlated inputs make the errors in different columns interact, which is what solvers such as GPTQ exploit. The outputs of a layer do not matter equally to the loss either: the second moment $G$ of the loss gradients at the outputs records which output directions matter, and it adds an interaction between rows in the Kronecker-factored form $\tr(GEHE^\top)$ \cite{martens2015kfac,tseng2026yaqa}. When such proxies are not accurate enough, the price of an edit can be measured directly, by running the network with the edit applied.

\subsection{One choice per matrix, one budget for the network}

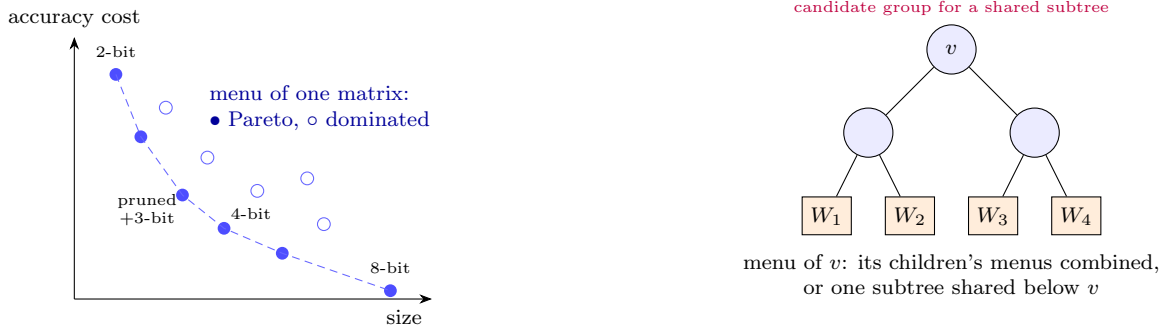
\begin{figure}[t]
\centering
\begin{tikzpicture}[x=0.55cm,y=0.55cm]
\draw[-{Stealth}] (0,0) -- (8.6,0) node[below left,font=\scriptsize] {size};
\draw[-{Stealth}] (0,0) -- (0,6.3) node[above,font=\scriptsize] {accuracy cost};
\foreach \x/\y in {1/5.4,1.6/3.9,2.6/2.5,3.6/1.7,5/1.1,7.6/0.2} {\fill[blue!70] (\x,\y) circle (2.4pt);}
\foreach \x/\y in {2.2/4.6,3.2/3.4,4.4/2.6,6/1.8,5.6/2.9} {\draw[blue!60] (\x,\y) circle (2.4pt);}
\draw[blue!60, densely dashed] (1,5.4) -- (1.6,3.9) -- (2.6,2.5) -- (3.6,1.7) -- (5,1.1) -- (7.6,0.2);
\node[font=\scriptsize, blue!60!black, align=left] at (5.9,4.6) {menu of one matrix:\\$\bullet$ Pareto, $\circ$ dominated};
\node[font=\tiny] at (1,5.95) {2-bit};
\node[font=\tiny] at (4.25,2.05) {4-bit};
\node[font=\tiny] at (7.6,0.75) {8-bit};
\node[font=\tiny, align=center] at (1.75,2.15) {pruned\\+3-bit};
\begin{scope}[xshift=9.4cm]
\node[inode] (q) at (4,6) {$v$};
\node[inode] (p1) at (2,4) {};
\node[inode] (p2) at (6,4) {};
\node[leaf] (m1) at (1,2) {$W_1$};
\node[leaf] (m2) at (3,2) {$W_2$};
\node[leaf] (m3) at (5,2) {$W_3$};
\node[leaf] (m4) at (7,2) {$W_4$};
\draw (q) -- (p1); \draw (q) -- (p2); \draw (p1) -- (m1); \draw (p1) -- (m2); \draw (p2) -- (m3); \draw (p2) -- (m4);
\node[font=\scriptsize, align=center] at (4,0.6) {menu of $v$: its children's menus combined,\\or one subtree
shared below $v$};
\node[font=\tiny, purple, align=center] at (4,7.0) {candidate group for a shared subtree};
\end{scope}
\end{tikzpicture}
\caption{Left: the menu of one matrix in the (size, accuracy cost) plane; only its Pareto set enters the optimizer (\cref{prop:mckp}). Right: nested candidate groups for shared subtrees; every group offers either its children's combined menus or one subtree shared by all of its matrices, and keeping Pareto sets at every node keeps the search exact. Points are illustrative.}
\label{fig:opt}
\end{figure}

With every method expressed as an edit and every edit priced, compressing a network becomes a single problem. Each matrix receives a menu of edits, covering all methods and all settings, each with a size and an accuracy cost, and the task is to choose one edit per matrix so that the total size fits a budget (\cref{fig:opt}).

\begin{property}[One optimizer]\label{prop:mckp}
If the accuracy costs of different matrices add up, the best choice is found by dynamic programming over the Pareto sets of the menus, exactly up to the resolution of the size grid, and it is never worse than applying any single method, or any uniform setting, that fits the same budget. When the candidate groups for shared subtrees are nested or disjoint, a recursion over the groups keeps the search exact (\cref{thm:a-mckp,thm:a-forest}).
\end{property}

Restricted to one family, this is the per-layer allocation that SPDY solves by dynamic programming \cite{frantar2022spdy} and HAWQ-V3 by integer programming \cite{yao2021hawqv3}; what the tree adds is that one menu can hold edits of every family. The guarantee holds with respect to the cost model. In practice, the costs of edits in different matrices are not exactly additive, and aggressive edits interact. A simple procedure keeps the exact search honest: build and measure the chosen edits together, update their costs, and solve again until the choice stops changing; \cref{app:exp} describes the version used in our experiments.

Finally, the budget need not be counted in bytes. Additions, latency or energy can take its place without changing the algorithm, which turns the optimizer into a compiler for a given hardware target. For state trees the most natural of these budgets is the number of additions, and it leads to the second reading of the tree: as a circuit.
\section{From trees to circuits}\label{sec:circuits}

\Cref{fig:example}c computed two weighted inputs with three additions and no multiplication. This section scales that picture up: first to a layer, then to a whole network, and finally to a first measurement on real weights. The motivation is simple. In digital logic a multiplier takes considerably more area and energy than an adder of the same width \cite{horowitz2014energy}, and a state tree whose states are powers of two needs no multiplier for its weights at all.

\subsection{A layer becomes wires and adders}

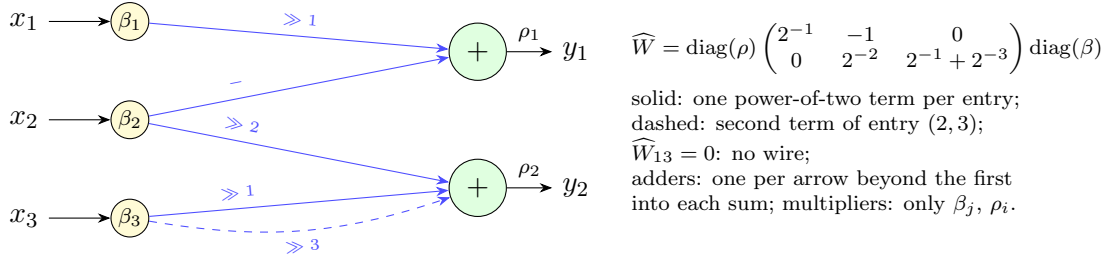
\begin{figure}[t]
\centering
\begin{tikzpicture}[>=Stealth]
\node (x1) at (0,0) {$x_1$}; \node (x2) at (0,-1.3) {$x_2$}; \node (x3) at (0,-2.6) {$x_3$};
\foreach \i/\y in {1/0,2/-1.3,3/-2.6} {
  \node[draw, circle, inner sep=1pt, font=\scriptsize, fill=yellow!20] (b\i) at (1.4,\y) {$\beta_{\i}$};
  \draw[->] (x\i) -- (b\i);}
\node[draw, circle, fill=green!12, minimum size=7mm] (s1) at (6,-0.4) {$+$};
\node[draw, circle, fill=green!12, minimum size=7mm] (s2) at (6,-2.2) {$+$};
\node[right=6mm of s1] (y1) {$y_1$}; \node[right=6mm of s2] (y2) {$y_2$};
\draw[->] (s1) -- node[above, font=\scriptsize] {$\rho_1$} (y1); \draw[->] (s2) -- node[above, font=\scriptsize] {$\rho_2$} (y2);
\draw[->, blue!70] (b1) -- node[above, sloped, font=\tiny, pos=0.5] {$\gg1$} (s1);
\draw[->, blue!70] (b2) -- node[above, sloped, font=\tiny, pos=0.3] {$-$} (s1);
\draw[->, blue!70] (b2) -- node[above, sloped, font=\tiny, pos=0.3] {$\gg2$} (s2);
\draw[->, blue!70] (b3) -- node[above, sloped, font=\tiny, pos=0.3] {$\gg1$} (s2);
\draw[->, blue!70, dashed] (b3) to[bend right=14] node[below, sloped, font=\tiny, pos=0.5] {$\gg3$} (s2);
\node[font=\scriptsize, align=left, anchor=west] at (7.9,-1.3) {%
  $\What=\diag(\rho)\begin{pmatrix}2^{-1}&-1&0\\0&2^{-2}&2^{-1}+2^{-3}\end{pmatrix}\diag(\beta)$\\[5pt]
  solid: one power-of-two term per entry;\\
  dashed: second term of entry $(2,3)$;\\
  $\What_{13}=0$: no wire;\\
  adders: one per arrow beyond the first\\
  into each sum; multipliers: only $\beta_j$, $\rho_i$.};
\end{tikzpicture}
\caption{A layer as a circuit. When the states are powers of two, multiplications by weights become shifts and signs, which are wiring; zeros are missing wires; and each additional power-of-two term costs one adder. The only multiplications left are by the column bases and the row scales.}
\label{fig:circuit}
\end{figure}

Suppose every state of a tree is a power of two, a short sum of powers of two, or zero. Then \cref{eq:value} turns the product $y=\What x$ into
\begin{equation}\label{eq:circuit}
  y_i\;=\;\rho_i\sum_{j=1}^{n}\ \sum_{\tau=1}^{t_{ij}}\ \pm\big((\beta_jx_j)\gg k_{ij\tau}\big),
\end{equation}
where $t_{ij}$ is the number of power-of-two terms in entry $(i,j)$ and, as in \cref{fig:example}, $\gg k$ shifts a number by $k$ binary positions, that is, multiplies it by $2^{-k}$ (a negative $k$ shifts to the left). In a circuit whose weights are fixed, \cref{eq:circuit} is a network of wires and adders (\cref{fig:circuit}). A shift is a re-labelling of wires and a sign is an inverted adder input, so both are free; a zero state is simply a missing wire; and every term beyond the first in
an output costs one adder. The only genuine multiplications left are the $n$ by the column bases and the $m$ by the row scales of an $m\times n$ matrix. They can be folded into neighbouring layers where no residual addition or nonlinearity intervenes, and otherwise remain; \cref{sec:adders} charges them in full.

\subsection{Where the circuits pay off}

The gain is largest where weights are fixed at fabrication: printed and bespoke classifiers \cite{mubarik2020printed}, and chips that hardwire an entire language model into silicon \cite{taalas2026hc1}. There, a weight with few nonzero digits is literally a few wires, and the tree states which ones. It is worth being precise about the source of this gain. A general multiplier is itself a shift-and-add circuit: Booth recoding writes one operand in signed digits and adds shifted copies of the other. A state tree makes the same structure explicit and tailors it to the actual weights: in a circuit with fixed weights, zero digits disappear entirely, and the tree can choose weights with few nonzero digits wherever accuracy allows. The same structure serves other targets as well. In programmable hardware, stored states select shifts and signs through small multiplexers, which take area of their own, so the saving over a general multiplier is smaller there, although a shift stored at a node is selected once for its whole group rather than for every weight, so deeper trees also save multiplexers; in software, lookup-table kernels \cite{park2024lutgemm,wei2024tmac} replace multiplications by table reads. Throughout, we assume fixed-point activations. With floating-point activations, adders must first align exponents, which remains cheap when the activations share an exponent per block, as in the microscaling formats.

\subsection{A whole network}

\Cref{tab:layers} goes through the components of common networks. Fully connected layers are exactly \cref{eq:circuit}, with biases as constant adder inputs. A convolution applies one kernel matrix at every position \cite{chellapilla2006conv}, so it is one shared subtree referenced at every position; its circuit is either replicated per position or built once and reused over time. Fast convolution algorithms are products with fixed transforms, and for Winograd's $F(2\times2,3\times3)$ \cite{lavin2016winograd} all three transforms have entries in $\{0,\pm\frac12,\pm1\}$, so they are shift trees themselves. The four projections of attention \cite{vaswani2017attention} are fully connected layers, and the query, key and value projections read the same input, so they can share partial sums. Batch normalization folds into neighbouring layers at inference \cite{jacob2018quantization}, and the gains of layer and RMS normalization fold into the column bases of the next layer. What remains outside the tree is small and well delimited: products of two activations, such as attention scores and the gate of a gated MLP, which are not weights and still need multipliers; and a few nonlinear functions, for which integer-only approximations are well studied \cite{kim2021ibert}.

\begin{table}[t]
\centering
\small
\renewcommand{\arraystretch}{1.15}
\begin{tabularx}{\textwidth}{@{}L{3.3cm}L{4.4cm}Y@{}}
\toprule
\textbf{Component} & \textbf{As a state tree} & \textbf{In a circuit}\\
\midrule
Fully connected $Wx+b$ & a tree (or graph) for $W$ & shifts and adders; the bias is a constant input\\
Convolution & one shared subtree, referenced per position & replicated per position, or reused over time\\
Winograd / fast convolution & products with fixed shift-tree transforms & shifts and adders only\\
Attention projections & four trees; $Q,K,V$ share their input & shared partial sums across $W_Q,W_K,W_V$\\
Attention scores, context & products of two activations & multipliers, or a separate treatment\\
Gated MLP & two trees that share their input & shared partial sums; a multiplier for the gate\\
Softmax, GELU, SiLU & --- & small tables or piecewise-linear units\\
Embedding & rows of a tree, decoded on lookup & memory only\\
Output head & a tree, possibly tied to the embedding & shifts and adders\\
Batch normalization & folded into $\rho$, $\beta$ and the bias & nothing left\\
Layer and RMS normalization & gain folded into the next $\beta$ & squares and a reciprocal square root per token\\
\bottomrule
\end{tabularx}
\caption{How the components of common networks map onto state trees and circuits. Every product of a weight and an activation becomes shifts and additions; what remains are products of two activations and a few nonlinear functions.}
\label{tab:layers}
\end{table}

\subsection{Sharing additions, and their limit}

Different outputs often need the same partial sum of shifted inputs, and can then share the adder that forms it. Finding such shared sums is the constant matrix--vector multiplication problem, a generalization of multiple constant multiplication \cite{voronenko2007mcm}. Entries below a common node already share their shifts, so the tree is a natural starting point for this search. There is also a limit, and it mirrors the one for storage in \cref{sec:express}. For matrices without structure, any circuit of additions and bounded shifts needs on the order of $n^2b/\log(nb)$ operations for most $n\times n$ matrices of $b$-bit integers, and, when $b$ is not exponentially larger than $n$, a classical bucketing construction comes within a constant factor of this \cite{arlazarov1970,liberty2009mailman} (\cref{prop:a-lower}). Large savings in additions, as in bits, come from structure, and the tree is where structure is recorded.

\subsection{A first measurement}\label{sec:adders}

Does choosing states for circuit cost pay off on real weights? We compared state trees with GPTQ \cite{frantar2023gptq} on five linear layers of LLaMA-3-8B, using 128 calibration sequences from FineWeb. GPTQ is designed for storage at a fixed bit width, not for additions, so the comparison asks how many additions a tree saves when it is allowed to target them. We use the flattest tree on purpose: a \emph{flat signed-digit tree} has one scale per row and leaves that are short sums of signed powers of two. This isolates the circuit price; deeper trees mainly pay off in bits. Both methods are costed by the same rule, one adder per nonzero signed digit of a weight and four adders per real-valued scale, which counts one adder per output more than the circuit of \cref{eq:circuit}, equally for both. For GPTQ the rule is applied to its quantized weights and group scales. For the tree, we choose the states to minimize the layer error $f_H$ of \cref{eq:fH} plus a price for every nonzero digit, and adjust the price until the layer error equals that of GPTQ at 4, 3 and 2 bits. The states are chosen in a single GPTQ-style pass: column by column, every weight becomes zero or one to three terms $\pm2^e$ times its row scale, with $e$ from $1$ down to $-11$, whichever adds least to the layer error plus the price of its digits, and the remaining columns receive the usual error update. The choice separates over the weights of a column, so each column is solved exactly given the columns before it. The row scale starts at the largest magnitude in its row and is refitted by least squares in the metric of $H$ after the pass. The weights of the tree are hardwired, so its storage is not counted. GPTQ is run in its plain form, with an asymmetric min--max grid for every group of 128 weights, no clipping or grid search, natural column order and 1\% damping; its digits are counted in the non-adjacent form of the signed code $q-z$.

\begin{table}[t]
\centering
\small
\begin{tabular}{@{}lccc@{}}
\toprule
Matrix & at GPTQ-4 error & at GPTQ-3 error & at GPTQ-2 error\\
\midrule
layer 1, gate projection  & 1.17 $\to$ 1.05 ($-10\%$) & 0.80 $\to$ 0.69 ($-14\%$) & 0.42 $\to$ 0.29 ($-30\%$)\\
layer 15, gate projection & 1.16 $\to$ 1.05 ($-9\%$)  & 0.80 $\to$ 0.69 ($-14\%$) & 0.42 $\to$ 0.29 ($-30\%$)\\
layer 15, down projection & 1.16 $\to$ 1.05 ($-10\%$) & 0.79 $\to$ 0.67 ($-15\%$) & 0.41 $\to$ 0.27 ($-35\%$)\\
layer 30, down projection & 1.15 $\to$ 0.93 ($-20\%$) & 0.79 $\to$ 0.56 ($-29\%$) & 0.40 $\to$ 0.16 ($-61\%$)\\
layer 1, down projection  & 1.16 $\to$ 0.79 ($-32\%$) & 0.80 $\to$ 0.36 ($-55\%$) & 0.41 $\to$ 0.15 ($-63\%$)\\
\bottomrule
\end{tabular}
\caption{Adders per weight at equal layer error on LLaMA-3-8B: GPTQ $\to$ a flat state tree whose
states were chosen for circuit cost. The last two matrices have strong outliers. Counts are rounded
to two decimals; percentages are computed from the unrounded counts.}
\label{tab:adders}
\end{table}

\Cref{tab:adders} shows the result. At equal layer error, the tree needs 9--15\% fewer adders than GPTQ at 3- and 4-bit accuracy on the three typical matrices, and 30--35\% fewer at 2-bit accuracy, a regime in which GPTQ itself already degrades the model severely, so that the 3- and 4-bit columns are the practical operating points. The gains are largest on the two matrices with strong outliers, up to 55\% at 3-bit and 63\% at 2-bit accuracy on the down projection of the first layer. Part of the saving comes from the scales, since GPTQ needs a scale for every group of 128 weights and the tree one per row. Counting weight digits alone, the tree still needs 6--7\% fewer adders on the typical matrices at 4-bit accuracy, 9--11\% at 3-bit and 23--29\% at 2-bit accuracy, and up to 53\% and 59\% on the outlier matrices at 3- and 2-bit accuracy. We attribute this to where the digits go: a uniform grid gives every weight the same number of digits, whereas the tree spends digits where they reduce the error most. A second observation concerns the shape of the tree, and it confirms what the two-weight example of \cref{sec:trees} suggested. In our trials, the flattest tree was also the best one for additions: the shared exponents of deeper trees saved stored bits but not additions, because a shift costs nothing in a circuit with fixed weights. Bits and additions are two different prices, and the representation makes both visible, so that the optimizer of \cref{sec:compress} can trade one against the other. Counting additions leaves out their width, which grows with the spread of the shifts that an adder combines, and the truncation of low-order bits by right shifts in fixed point. The counts are therefore a first-order proxy for circuit size, not a measurement of area. The flat tree also differs from earlier multiplier-free quantizers. APoT \cite{li2020apot} draws its levels from a fixed additive power-of-two grid shared by a whole layer and learned with quantization-aware training, and ShiftAddLLM \cite{you2024shiftaddllm} writes weights as sums of binary matrices with power-of-two group scales; the flat tree chooses the number and position of the signed digits separately for every weight and prices each digit against the layer error, so a zero costs nothing and an outlier costs one digit. A full study of circuits for whole networks, including area, energy, these effects and measured comparisons with quantizers that also target additions \cite{li2020apot,faraone2020addnet,you2024shiftaddllm}, is left to a separate paper.
\section{Choosing states is finding a ground state}\label{sec:bm}

In the previous two sections the states of a tree were chosen by some solver, either for accuracy at a given size or for accuracy at a given number of additions. It is worth asking what kind of problem these solvers face. The answer is precise and well studied: choosing the states of a tree is the problem of finding the lowest-energy configuration of a Boltzmann machine.

\subsection{Boltzmann machines}

A Boltzmann machine \cite{ackley1985boltzmann} consists of $N$ binary units $z_1,\dots,z_N$, each either off ($z_p=0$) or on ($z_p=1$); a \emph{configuration} $z=(z_1,\dots,z_N)$ fixes the state of every unit. Each configuration has an energy \begin{equation}\label{eq:bm} \mathcal E(z)\;=\;\sum_{p<q}J_{pq}\,z_pz_q+\sum_ph_p\,z_p , \end{equation} where the indices $p$ and $q$ run over the units. The \emph{coupling} $J_{pq}$ is the energy added when units $p$ and $q$ are on together: a positive coupling discourages them from being on at the same time, and a negative one encourages it. The \emph{field} $h_p$ is the energy added when unit $p$ is on by itself. The machine assigns each configuration the probability $P(z)=\exp(-\mathcal E(z)/T)/Z$, where the \emph{temperature} $T>0$ sets how strongly low energies are preferred, and $Z$, the sum of $\exp(-\mathcal E(z)/T)$ over all configurations, makes the probabilities add up to one. As $T$ falls, the probability concentrates on the configurations of lowest energy, the \emph{ground states}. A higher-order machine \cite{sejnowski1986higher} also has terms that couple three or more units, such as $J_{pqr}\,z_pz_qz_r$. Finding a ground state of \cref{eq:bm} is the problem known as QUBO (quadratic unconstrained binary optimization); written with spins $\sigma_p=2z_p-1\in\{-1,+1\}$ instead of bits, it is the Ising model \cite{lucas2014ising}.

\subsection{A machine with two weights}

Return to the two weights of \cref{fig:example}, and suppose that their mantissas must be shortened, so that each weight has to choose between its two nearest representable values. Each choice is a binary unit: $z_1$ for $w_1$ and $z_2$ for $w_2$. The natural energy is the layer error of \cref{eq:fH}. For the two inputs $x_1,x_2$ it reads
\[
  f_H\;=\;e_1^2H_{11}\;+\;2\,e_1e_2\,H_{12}\;+\;e_2^2H_{22},
\]
where $e_1$ and $e_2$ are the errors of the two weights, each affine in its own unit. The first and last terms depend on one unit each. The middle term couples them, with a strength set by $H_{12}$, the correlation between the two inputs. If the inputs are uncorrelated, $H_{12}=0$ and each weight can be rounded on its own, which is exactly what rounding to the nearest value does. If they are correlated, the best choice for one weight depends on the choice made for the other: the two units are coupled, and the error is a small Boltzmann machine. Now let the shared node take part as well, with a bit $b$ that either keeps or halves both weights. Each weight is then the product of the node's factor and its leaf, so the error contains products of $b$ with $z_1$ and $z_2$: terms that couple three units at once. These higher-order terms follow the path from the shared node to its leaves. These two observations, couplings from the data and higher-order terms from the tree, are all there is to the general case.

\subsection{The Boltzmann machine of a tree}

Every state of a tree can be written with binary units. A leaf with $K$ possible values becomes a group of $K$ units of which exactly one is on. A shift bit with multiplier $r^b$ is affine in its bit, $r^b=1+(r-1)b$. A gate multiplies its subtree by a bit, and a lookup index is again a group of units of which exactly one is on. Every weight is then a polynomial in the units on its path, and the layer error becomes an energy. One-hot groups need a penalty that keeps exactly one unit on, but many alphabets need none: a leaf $\{0,\pm1\}$ is $z-z'$ with two free units, zero having two codes, which leaves the energy flat between them and does no harm to the search, and an integer range has a binary encoding that is linear in its units.

\begin{property}[The Boltzmann machine of a tree]\label{prop:bm}
With the node states fixed, the layer error is a quadratic function of the leaf units: a pairwise Boltzmann machine in which leaves of the same row are coupled through the input statistics $H$, and, under the Kronecker-factored error, leaves of different rows are also coupled through the output statistics $G$. With the node states free as well, the error is a higher-order Boltzmann machine whose higher-order terms follow the paths of the tree; auxiliary units reduce it to a pairwise machine \cite{boros2002pseudo,ishikawa2011transformation}. In both cases, for given row scales and column bases, the ground state is the best matrix the tree can represent (\cref{thm:a-bm}).
\end{property}

\begin{figure}[t]
\centering
\begin{tikzpicture}[>=Stealth]
\tikzset{unit/.style={draw, rounded corners=2pt, fill=orange!14, minimum width=11mm, minimum height=5mm, font=\scriptsize}}
\foreach \q/\x in {1/0,2/1.7,3/3.4,4/5.1} {
  \node[unit] (j\q) at (\x,0) {$z_{\q j}$};
  \node[unit] (k\q) at (\x,-1.7) {$z_{\q k}$};}
\foreach \q in {1,...,4} {\draw[blue!70, thick] (j\q.south) -- (k\q.north);}
\node[font=\tiny, blue!70!black, fill=white, inner sep=1pt] at (0,-0.85) {$H_{jk}$};
\node[inode] (aj) at (0.85,1.15) {$b$};
\node[inode] (cj) at (4.25,1.15) {$b$};
\node[inode] (rj) at (2.55,2.05) {$b$};
\draw[black!55] (aj) -- (j1); \draw[black!55] (aj) -- (j2); \draw[black!55] (cj) -- (j3); \draw[black!55] (cj) -- (j4);
\draw[black!55] (rj) -- (aj); \draw[black!55] (rj) -- (cj);
\node[inode] (ak) at (0.85,-2.85) {$b$};
\node[inode] (ck) at (4.25,-2.85) {$b$};
\node[inode] (rk) at (2.55,-3.75) {$b$};
\draw[black!55] (ak) -- (k1); \draw[black!55] (ak) -- (k2); \draw[black!55] (ck) -- (k3); \draw[black!55] (ck) -- (k4);
\draw[black!55] (rk) -- (ak); \draw[black!55] (rk) -- (ck);
\begin{scope}[on background layer]
\draw[line width=7pt, purple!18, line cap=round, line join=round]
  (rj.center) -- (aj.center) -- (j1.center) -- (k1.center) -- (ak.center) -- (rk.center);
\end{scope}
\draw[red!70, thick, dashed] (j2.south east) -- (k3.north west);
\node[font=\tiny, red!70!black, fill=white, inner sep=1pt] at (2.55,-0.85) {$G_{23}H_{jk}$};
\node[font=\scriptsize, anchor=east] at (-0.75,0) {column $j$};
\node[font=\scriptsize, anchor=east] at (-0.75,-1.7) {column $k$};
\node[font=\scriptsize, anchor=west] at (6.3,-0.85) {%
  \renewcommand{\arraystretch}{1.1}%
  \begin{tabular}{@{}l@{\ \ }l@{}}
  orange & leaf units (rows 1--4 of two columns)\\
  circles & shared shift bits of each column\\
  shaded & one higher-order term: the units\\
         & on the paths to $(1,j)$ and $(1,k)$\\
  blue & same row, two columns: from $H$\\
  red dashed & two rows: from $G$ (two-sided error)
  \end{tabular}};
\end{tikzpicture}
\caption{The Boltzmann machine of a state tree (\cref{prop:bm}), drawn for two columns of a four-row version of the running example. The tree decides which units exist and where the higher-order terms are; the data decide how strongly the units interact.}
\label{fig:bm}
\end{figure}

\Cref{fig:bm} shows the same division of labour as the two-weight machine, now for a larger tree: the tree decides which units exist and where the higher-order terms are, and the data decide how strongly the units interact.

\subsection{Familiar solvers, and new ones}

Seen through this machine, familiar solvers are particular ways of lowering its energy. Rounding to the nearest value ignores all couplings, as in the uncorrelated case above. GPTQ visits the columns in turn and fixes each one given the columns already fixed, carrying their errors forward to the columns that remain; it is a single sweep at zero temperature that never revisits a decision, and, run in reverse column order and without clipping, it coincides with Babai's nearest-plane algorithm for lattices \cite{babai1986,chen2026gptqbabai}. Adaptive rounding writes the rounding of a layer as a QUBO \cite{nagel2020adaround}; in the present language, this is a flat tree whose leaves choose between rounding up and rounding down, exactly the two-weight machine without its shared node. 

The machine also suggests solvers that are rarely used for this problem: Gibbs sampling, simulated annealing \cite{kirkpatrick1983annealing}, parallel tempering, and sampling at a finite temperature to obtain many good solutions instead of one. Because the energy covers every kind of decision, including zeros, shared shifts, codebook indices and selectors, the same solver can choose all of them together.

\subsection{Quantum optimizers}

Quantum annealers \cite{kadowaki1998quantum} and the quantum approximate optimization algorithm \cite{farhi2014qaoa} search for low-energy states of Ising models. After the substitution $\sigma=2z-1$ and, when node states are free, a standard reduction to pairwise terms, the energy of a tree has exactly this form; running it on a device still requires an embedding of the couplings and their representation at finite precision. A whole weight matrix has far more units than present devices have qubits, but the tree also supplies a decomposition: one group of rows and one block of columns at a time, with GPTQ-style error feedback carrying the effect of each solved block to the rest. Each block is a dense machine of tens to a few hundred units, the size handled by current annealers and small variational circuits, inside a classical loop that walks the tree. These sizes hold when the node states of a block are fixed, so that its machine is pairwise from the start; freeing them adds auxiliary units and penalty couplings, which enlarge the machine and widen the range of couplings a device must represent. A single large matrix splits into a very large number of such blocks, and whether quantum solvers will outperform classical ones on them is an open question. What the representation provides is the problem in the form these devices accept, together with a decomposition that matches their size.

The same tree helps in the opposite direction, when a network has to be evaluated inside a quantum algorithm. There the weights are classical constants, and multiplying a quantum register by a constant is compiled into shifted additions, whose cost grows with the number of nonzero digits of the constant, while a shift is only a relabelling of qubits \cite{cuccaro2004adder}. A state tree controls exactly this number.

Finally, the temperature of the machine points to a different use of the tree: not only to choose the states of a network that has already been trained, but to learn them in the first place. This is the subject of the next section.
\section{Learning in the space of states}\label{sec:learning}

So far a tree was fitted to a network trained in the usual way. The representation also invites training the tree directly. Ternary networks trained from scratch \cite{ma2024era,ma2025bitnet2b4t} already do this for the simplest tree, one scale above leaves $\{-1,0,1\}$, and quantization-aware training does it for group-quantization trees. Deeper trees can be trained in the same way, through a relaxation of the discrete choices: every alphabet receives a soft distribution over its states, which is sharpened as a temperature is lowered. Training then moves through the space of decisions, and the trained object is already the deployed one: its storage cost, its circuit and its format are fixed by the tree. The temperature plays the same role as in the Boltzmann machine of \cref{sec:bm}, and connecting the two precisely for training losses is a natural direction for future work.

\section{Early evidence and directions}\label{sec:directions}
\providecommand{\impact}{\par\smallskip\noindent\textbf{\textit{Impact.}}\ }

The previous sections developed one idea from several sides. This section shows that it already delivers, on real models and beyond them (\cref{tab:evidence,tab:images}). We present five results, each followed by what it means in practice. They are selected examples rather than a general claim, and each will be reported in full in a separate paper; together, they show that a single representation brings concrete gains in places as different as model serving, compression, model tuning, hardware design and image coding.

\begin{table}[t]
\centering
\footnotesize
\renewcommand{\arraystretch}{1.12}
\begin{tabularx}{\textwidth}{@{}L{2.3cm}Y L{2.7cm}L{3.3cm}@{}}
\toprule
\textbf{Reading} & \textbf{Setting and metric} & \textbf{Baseline} & \textbf{State tree}\\
\midrule
Storage and\newline serving & BF16 weights stored exactly; bits per weight & BF16: 16 bits &
\textbf{12.13 bits ($-24\%$), bit-exact}\\

 & LLaMA-3-8B, layer 15; matrix--vector speed-up, batch 1--8 & cuBLAS (A100);\newline PyTorch (CPU)
& \textbf{1.05--1.51$\times$; 1.24--1.27$\times$}\\
 & same matrices and batches, format alone (bound $1.32\times$) & our kernel on BF16 (A100)
& \textbf{1.03--1.27$\times$}\\
 & vLLM end-to-end serving, Llama-3.2-1B and Qwen3.5-9B; speed-up at batch 1 / 8 / 32 & vLLM,
BF16 & \textbf{1.23$\times$ / 1.17--1.21$\times$ / $\approx$1.0$\times$}\\
\hline
Compression & LLaMA-3-8B at 3.25 bits per weight, no training; perplexity on WikiText-2 / C4 /
FineWeb & strongest GPTQ:\newline 8.09 / 12.72 / 11.81 & \textbf{6.95 / 10.93 / 9.76}\\
\hline
Discrete\newline optimization & Qwen2.5-1.5B-Instruct at 2.44 bits per weight, only states change;
validation KL, GSM8K, MMLU & before tuning:\newline 0.491, 13.95, 48.3 & \textbf{0.371, 21.83,
49.6}\\
 & LLaMA-3-8B, layer 1 down projection, 3.3-bit tree; relative output error & before sampling: 0.0267 &
\textbf{0.0153}\\
\hline
Circuits & LLaMA-3-8B, adders per weight at equal layer error (\cref{tab:adders}) & GPTQ at 4 / 3 / 2
bits & \textbf{9--32\% / 14--55\% / 30--63\% fewer}\\
\bottomrule
\end{tabularx}
\caption{One result for each reading of the tree. The strongest GPTQ uses act-order and an MSE-optimized grid in the same pipeline. Perplexity at context 2048 (FP16: 6.14 / 9.44 / 7.85); bits per weight cover all linear layers of the transformer blocks, scales included.}
\label{tab:evidence}
\end{table}

\paragraph{Lossless storage: less memory and faster serving, for free.}
The simplest use of the tree involves no approximation at all. \Cref{sec:trees} showed that two neighbouring weights can share their exponent and still be reproduced exactly. At scale, a BF16 matrix stored with one shared exponent per 64 weights and, per weight, a 4-bit state and an 8-bit leaf needs 12.13 instead of 16 bits per weight over whole models, and every weight decodes to exactly its BF16 value. Weights whose exponent lies 15 or more below the largest in their group, about 0.01\% of them and including exact zeros, are kept exactly in a short side list, whose cost is included (\cref{app:exp}). Fewer bytes matter because generating tokens is limited by memory traffic rather than by arithmetic. Our CUDA kernel decodes the tree with byte-parallel integer operations, feeds the tensor cores directly and runs at 97\% of the bandwidth floor of the tree bytes. Reading 12.13 instead of 16 bits per weight bounds the gain from the format alone at $16/12.13\approx1.32\times$. \Cref{tab:gemv} separates the two sources of speed on layer~15 of LLaMA-3-8B, with every kernel timed in the same run. Against our own kernel reading plain BF16, the tree makes matrix--vector products 1.03--1.27$\times$ faster at batch sizes 1 to 8, below the bound; this is the gain of the format. Against cuBLAS the gain is 1.05--1.51$\times$, because on the large MLP matrices our kernel design is itself up to 1.27$\times$ faster than cuBLAS, while on the attention projections cuBLAS is already as fast and the whole gain comes from the format. At batch 32 the product becomes compute-bound, and the tree is up to 15\% slower than our BF16 kernel. On a CPU, products become 1.24--1.27$\times$ faster than PyTorch. In end-to-end serving with vLLM, generation becomes 1.23$\times$ faster at batch size 1 for both Llama-3.2-1B \cite{llama3} and Qwen3.5-9B \cite{qwen35}, and 1.17--1.21$\times$ faster at batch size 8. When serving becomes compute-bound, at batch size 32, the tree runs at a speed comparable to BF16. These end-to-end gains include both the format and our kernel design; separating them requires serving the same model with our kernel reading BF16, which we leave to the companion paper. \impact A BF16 model can be stored with 24\% less disk and served with 24\% less weight memory and up to 23\% higher throughput at the small batch sizes of latency-sensitive serving, with tiny slowdown in the compute-bound regime. Because the weights are exactly the same, all of this comes without any loss of accuracy, calibration or retraining, and the memory it frees can hold a longer key--value cache or a larger batch. The vLLM plugin is enabled with a single flag and is publicly available.\footnote{Code and vLLM plugin: \url{https://github.com/ammarali32/StateTree-Representation}.}
\begin{table}[h]
\centering
\small
\resizebox{\textwidth}{!}{
\begin{tabular}{@{}lrrrrccc@{}}
\toprule
Matrix & Batch & cuBLAS & own BF16 & tree & kernel & format & total\\
 & & ($\mu$s) & ($\mu$s) & ($\mu$s) & {\scriptsize own BF16 vs cuBLAS} & {\scriptsize tree vs own BF16} & {\scriptsize tree vs cuBLAS}\\
\midrule
gate $14336\times4096$ & 1  & 106.0 & 85.6  & 70.3  & 1.24$\times$ & 1.22$\times$ & 1.51$\times$\\
                       & 8  & 107.8 & 84.7  & 72.4  & 1.27$\times$ & 1.17$\times$ & 1.49$\times$\\
                       & 16 & 108.9 & 86.6  & 78.4  & 1.26$\times$ & 1.10$\times$ & 1.39$\times$\\
                       & 32 & 117.2 & 98.3  & 113.7 & 1.19$\times$ & 0.86$\times$ & 1.03$\times$\\
down $4096\times14336$ & 1  & 102.8 & 89.8  & 70.9  & 1.14$\times$ & 1.27$\times$ & 1.45$\times$\\
                       & 8  & 103.0 & 89.9  & 73.3  & 1.15$\times$ & 1.23$\times$ & 1.41$\times$\\
                       & 32 & 111.7 & 101.4 & 118.6 & 1.10$\times$ & 0.85$\times$ & 0.94$\times$\\
q $4096\times4096$     & 1  & 28.9  & 28.2  & 24.3  & 1.02$\times$ & 1.16$\times$ & 1.19$\times$\\
                       & 8  & 29.3  & 28.8  & 26.3  & 1.02$\times$ & 1.10$\times$ & 1.11$\times$\\
k $1024\times4096$     & 1  & 10.7  & 10.5  & 10.2  & 1.02$\times$ & 1.03$\times$ & 1.05$\times$\\
\bottomrule
\end{tabular}
}
\caption{Matrix--vector products of layer~15 of LLaMA-3-8B on an A100, all three kernels timed in
the same run. ``Own BF16'' is our kernel with the same structure reading plain BF16, so
\emph{kernel} isolates kernel engineering and \emph{format} isolates the 12.13-bit tree (bounded by
$16/12.13\approx1.32\times$); their product is the total gain over cuBLAS. At batch 32 the product
is compute-bound and decoding costs more than the saved bytes gain.}
\label{tab:gemv}
\end{table}
\paragraph{Compression: better low-bit models, without training.}
\Cref{sec:compress} argued that once every compression method is an edit of the same object, a single optimizer can choose among all of them for every matrix. A dynamic program over the whole of LLaMA-3-8B does exactly this: at 3.25 bits per weight and without any training, it lowers perplexity by 14--17\% relative to the strongest GPTQ configuration in the same pipeline, and closes 58\% of GPTQ's gap to FP16 on WikiText-2 (from 1.95 to 0.81). The accuracy cost of every option is measured on the calibration loss $L$, as $\frac12[L(W+\Delta)+L(W-\Delta)]-L(W)$ with $\Delta$ the error of the option, which removes the first-order term and puts all matrices and methods on one scale. The menu of every matrix combines quantization, pruning and low-rank edits, with no rotation and no trellis-coded quantizer, and will be described in a follow-up paper; every option the solver picks is built and measured, and the problem is solved again until the whole plan has been measured (\cref{app:exp}). Instead of one method being forced on every matrix, each matrix receives the edit that buys the most accuracy for its bits. \impact At a fifth of the size of FP16, the compressed model is much closer to the original than GPTQ at the same size, with no training at all. And because every method is just one more edit on the menu, a new compression method does not have to compete with the existing ones: it can be added to the optimizer and combined with all of them.

\paragraph{Discrete optimization: a better model in the same bytes.}
A compressed model written as a tree is a set of discrete states, and \cref{sec:bm} showed that choosing them is a ground-state problem. A model can therefore keep improving after compression without changing its format or its size. In \emph{Ising healing}, every step models the KL divergence to the original model by its gradient and its Kronecker factors, estimated on about 45k tokens, and gives every movable code one unit that either keeps it or moves it by one level within its own grid. The energy is then the Kronecker-factored machine of \cref{prop:bm} with an added linear term; it is solved with ballistic simulated bifurcation, a quantum-inspired classical solver \cite{goto2019sb,goto2021ballistic}, started from a greedy update, and a step is kept only if the KL on a validation set improves. As an illustration, we compressed Qwen2.5-1.5B-Instruct without training to 2.44 bits per weight with a wider menu, whose details will follow in a separate paper, and let only its GPTQ-family leaves, 6.8\% of the weights, move. Forty such steps changed 46,239 stored codes, about 0.004\% of the model's codes, and lowered the validation KL to the original model by 24\%; since this KL guided the search, we report it as a validation result only. On benchmarks that play no part in the search, healing improved perplexity on all three corpora (on WikiText-2 from 13.74 to 12.85) and raised GSM8K accuracy by 7.9 points, from 13.95 to 21.83. GSM8K has 1,319 problems, so each of these accuracies carries a 95\% binomial interval of about $\pm2$ points and the gain is about five standard errors; it comes from a single run. The solver reached lower energies than a greedy update on 371 of 516 real instances and higher ones on none, but end to end the choice of solver matters little. The reported run keeps, at every step, the best of no move, the greedy update and simulated bifurcation; run alone in a separate ablation, greedy updates reach a validation KL of 0.378 and GSM8K 20.3, and simulated bifurcation 0.371 and 21.0. With the same number of tokens, tuning only the continuous scales of the same model helps more (KL 0.325, GSM8K 27.5); healing changes the discrete codes instead, so the two act on different parameters. Annealed Gibbs sampling over the states of a 3.3-bit tree, six sweeps followed by two of coordinate descent, reduced the output error of the down projection of the first layer of LLaMA-3-8B by 43\%.
\impact A compressed model can be repaired after the fact with the same file size, the same format and the same kernels; the repair touches a few tens of thousands of codes and can be shipped as a tiny patch. The Boltzmann reading turns such repairs into a standard discrete problem, open to solvers from statistical physics today and to quantum annealers as they mature, although on this first instance simple solvers already capture most of the gain.

\paragraph{Hardware: fewer adders for the same accuracy.}
At equal layer error, choosing the states of a tree for circuit cost needs 14--55\% fewer adders than GPTQ at 3-bit accuracy and 30--63\% fewer at 2-bit accuracy, the upper ends on the two matrices with strong outliers (\cref{sec:adders}). The saving is structural: every weight carries its own power of two, so an outlier costs one adder instead of forcing extra precision on its whole group; zeros and shifts are free; and no scale multipliers per group remain.
\impact For chips that hardwire a model, the number of adders is a first-order measure of silicon area and energy. The tree lets a designer choose the weights for the circuit rather than for storage, and spend adders only where accuracy needs them.

\begin{table}[t]
\centering
\footnotesize
\renewcommand{\arraystretch}{1.12}
\begin{tabular}{@{}llcc@{}}
\toprule
\textbf{Codec} & \textbf{Implementation} & \textbf{Bits per pixel} & \textbf{Decoding (Gpixel/s)}\\
\midrule
JPEG XL, effort 9 \cite{alakuijala2019jpegxl} & libjxl 0.11.2, CPU & 8.66 & ---\\
\textbf{Ours (tree analogue)} & GPU (A100), batched & \textbf{9.47} & \textbf{2.9}\\
WebP, lossless \cite{rfc9649webp} & libwebp 1.6.0, CPU & 9.50 & ---\\
JPEG 2000 \cite{taubman2002jpeg2000} & OpenJPEG 2.5.4; GPU / CPU & 9.57 & 0.10 / $\approx$0.04\\
HTJ2K \cite{taubman2019htj2k} & OpenJPH 0.26.3; GPU decoding & 10.53 & 0.45\\
\bottomrule
\end{tabular}
\caption{Lossless image coding on Kodak \cite{kodak}; every result was checked to be bit-exact. GPU decoding of JPEG 2000 and HTJ2K uses nvImageCodec with the nvJPEG2000 backend. All GPU timings decode the 24 images in batched calls; one image per call, JPEG 2000 and HTJ2K reach 0.03 and 0.39 Gpixel/s. GPU timings run from compressed bytes in host memory to the decoded image in GPU memory (median of 10 runs, A100-SXM4-40GB); the CPU timing, on 16 cores of a shared AMD EPYC 7742, is indicative. For a single small image, GPU HTJ2K keeps an advantage that the batched timings do not show.}
\label{tab:images}
\end{table}

\paragraph{Beyond neural networks: fast lossless images.}
Nothing in the representation is specific to weights: any array of integers can be written as paths of decisions. We tested this on lossless image coding, using the same path that \cref{sec:trees} applied to a weight. After a reversible colour transform chosen per image, each pixel is predicted coarse to fine from four already decoded neighbours, in twelve passes per channel, each of which decodes fully in parallel. The prediction residual, mapped to a non-negative integer, is then written as a path of two decisions: its \emph{state} is its bit length, the counterpart of an exponent, and its \emph{leaf} holds the bits below the leading one, the counterpart of a mantissa. There is one twist. Instead of storing a shared node for a group of pixels, the decoder infers it: a context computed from the decoded neighbours, their range and their recent residual activity, plays the role of the node and selects the probability model under which the state and the first leaf bit are coded with interleaved rANS \cite{duda2013ans,giesen2014interleaved}; the remaining leaf bits are stored as they are. The shared decision thus becomes shared knowledge rather than shared storage. The codec is therefore an analogue of a state tree rather than an instance of \cref{def:a-tree}: its decisions are entropy-coded under inferred contexts instead of being stored as states. The probability models come from a small codebook trained once on 40 DIV2K images \cite{agustsson2017div2k}, disjoint from the test sets.

The result is a codec whose main strength is speed (\cref{tab:images}). On Kodak it needs 9.47 bits per pixel, less than lossless WebP (9.50), JPEG 2000 (9.57) and HTJ2K (10.53), though 9\% more than JPEG XL at its strongest setting (8.66); on the CLIC professional validation set \cite{toderici2020clic} it needs 8.51, against 7.88 for JPEG XL and 8.81 for lossless WebP (\cref{app:exp}). Decoding batched images on an A100 runs at 2.9 Gpixel/s on both datasets, against 0.45 for HTJ2K and 0.10 for JPEG 2000 on the same GPU, and about 0.04 for JPEG 2000 on sixteen CPU cores. For a single small image, GPU HTJ2K keeps an advantage that the batched timings do not show, and the strongest context-modelling coders, JPEG XL and paq8px, still compress photographs better. \impact Exact images decode about $6\times$ faster than with HTJ2K and almost $30\times$ faster than with JPEG 2000 on the same GPU, at a better compression ratio than both. Training pipelines in which data loading is the bottleneck, and the streaming of images and textures, gain directly. More broadly, the result shows that the representation is a property of numbers, not of neural networks.

\paragraph{What comes next.}
These results cover only part of what the representation opens, and each suggests a direction of its own.
\begin{itemize}[leftmargin=*]
\item \emph{Learning in tree space.} Training deep trees from scratch (\cref{sec:learning}) would let networks discover which decisions to share, instead of having the sharing imposed after training.
\item \emph{From model to silicon.} Compiling whole models into hardwired adder circuits would turn \cref{sec:circuits} into a design flow with explicit trade-offs between area, energy and accuracy.
\item \emph{Sparse model updates.} Since tuning moves only a tiny fraction of the states, updates and fine-tuning deltas can be shipped as small edits of a tree.
\item \emph{Beyond weights.} Activations and the key--value cache could be written as trees on the fly, and the image result suggests that other numerical data, such as scientific arrays, audio and video, can benefit from the same structure.
\item \emph{Quantum hardware.} The block subproblems of \cref{sec:bm} are natural first workloads for quantum annealers and variational circuits as they grow.
\end{itemize}
In every case the gain has the same source: the decisions that numbers share are made explicit, stored once, and optimized as discrete states.
\section{Related work}\label{sec:related}

TWIG brings together ideas from several lines of work that have so far developed separately.

\paragraph{Number formats with shared parts.}
Group quantization shares a scale, and block floating point and the microscaling formats share an exponent across a block of weights \cite{rouhani2020msfp,rouhani2023mx}. These are state trees with two levels of decisions (\cref{tab:formats}). Shared microexponents \cite{rouhani2023sharedmx} and HiFloat4 \cite{luo2026hif4} go deeper, with one or two further levels of one-bit shifts shared by sub-blocks, and HiFloat4 reports higher accuracy than NVFP4 on several language models at the same 4.5 bits per weight. TWIG allows sharing at any level and of any kind of decision, and makes its cost in bits and in additions explicit.

\paragraph{Path-structured representations.}
Describing a function or a matrix by paths through a graph with values on its edges, with identical parts stored once, is a classical idea in circuit verification and quantum computing. Edge values are additive in edge-valued decision diagrams \cite{lai1992evbdd} and multiplicative in binary moment diagrams \cite{bryant1995bmd} and in the decision diagrams that store the matrices of quantum circuits \cite{miller2006qmdd}; weighted automata use the same principle to compress images \cite{culik1993wfa}. In numerical linear algebra, hierarchical matrices store low-rank blocks over a tree of index sets \cite{hackbusch1999hmatrix}, and quantized tensor trains write an entry as a product of factors indexed by the binary digits of its position \cite{oseledets2011tt,khoromskij2011qtt}, close to the product along a path in \cref{eq:value}. Tree-structured and residual vector quantizers \cite{gersho1992vq}, and binary-coded quantization, which writes a matrix as a sum of scaled binary matrices \cite{xu2018amq,park2024lutgemm}, share decisions in related ways. TWIG adopts this path structure for the weights of neural networks. What is new is the cost model of bits and additions, the treatment of compression methods as edits of one object, and the reading of state selection as a Boltzmann machine.

\paragraph{Multiplier-free networks.}
Logarithmic and power-of-two weights \cite{miyashita2016log,zhou2017inq,elhoushi2021deepshift}, sums of power-of-two terms \cite{li2020apot}, and networks built from additions \cite{chen2020addernet} or from shifts and additions \cite{you2020shiftaddnet,you2024shiftaddllm} avoid multipliers by choosing a multiplier-free family of weights, during or after training. In the language of this paper these families are particular state trees, and \cref{prop:shift} explains why sums of power-of-two terms lose no expressive power. Signed-power-of-two coefficients and multiplierless constant multiplication have long been used in digital filter design, where the number of nonzero digits is a cost separate from the word length \cite{lim1999spt,voronenko2007mcm}. Reconfigurable constant-coefficient multipliers bring this to neural networks on FPGAs, restricting weights to coefficients that adders, shifts and multiplexers can apply \cite{faraone2020addnet}, and lookup-table kernels bring related ideas to processors \cite{park2024lutgemm,wei2024tmac}.

\paragraph{Unified and budgeted compression.}
Optimal Brain Compression treats pruning and quantization in one framework \cite{frantar2022obc}, SPDY chooses per-layer settings under a speed budget by dynamic programming \cite{frantar2022spdy}, and mixed-precision methods allocate bits per layer, for instance by integer programming \cite{yao2021hawqv3}, and ROCKET allocates low-rank truncation across layers as a knapsack problem
\cite{ali2026rocket}. TWIG supplies a common object for such choices, so that a single menu can contain edits from every family (\cref{sec:compress}).

\paragraph{Discrete optimization of weights.}
Adaptive rounding writes the rounding of a layer as a QUBO \cite{nagel2020adaround}; GPTQ has been identified with Babai's nearest-plane algorithm, under conditions on column order and clipping \cite{chen2026gptqbabai}; Kronecker-factored curvature has been used to round with both input and output statistics \cite{tseng2026yaqa}; and quantization and pruning have been formulated for simulated and quantum annealing \cite{munizsubinas2025qubo,wang2025adiabatic}. \Cref{prop:bm} extends the QUBO view to all decisions of a tree, including zeros, shared shifts, codebook indices and selectors.

\section{Conclusion}

Neural networks are trained in the continuum but run in a discrete world, and the passage between the two has so far been handled one method at a time. We have proposed to make this passage explicit with a single representation. In TWIG, every weight is the end of a path of small discrete decisions, and weights that belong together share the upper part of their paths, so that each shared decision is stored once and applied once. The idea is simple, and its consequences reach in several directions. The formats in use today are shallow state trees; nothing is lost in the worst case, and shared structure lowers the cost in bits. Read as a compressed model, a tree turns pruning, quantization, factorization, codebooks and sharing into edits of one object, chosen by one optimizer. Read as a circuit, it applies a layer with shifts and additions, apart from one scale per row and per column, with a size bounded by a count read off the tree. Read as an optimization problem, choosing its states is the ground-state problem of a higher-order Boltzmann machine whose interactions follow the paths of the tree, the problem class that annealers and quantum optimizers are built to solve. And because its decisions are explicit, a tree could be trained directly, so that the trained object is already the deployed one, a direction we leave open.

\appendix
\section{Definitions, statements and proofs}\label{app:formal}

Throughout, $W\in\R^{m\times n}$ is a weight matrix (rows are outputs, columns are inputs), $I=[m]\times[n]$ is the set of its entries, and $\Pset=\{0\}\cup\{\pm2^k:k\in\Z\}$ is the set of signed powers of two together with zero. Several statements below are known results restated for state trees: the counting bounds of \cref{thm:a-count,thm:a-density}, the non-adjacent form in \cref{thm:a-shift}, reduced-rank regression in \cref{thm:a-rank} \cite{izenman1975rrr}, the multiple-choice knapsack problem in \cref{thm:a-mckp} \cite{sinha1979mckp}, the bucketing construction in \cref{prop:a-lower}, and the quadratic form of rounding in \cref{thm:a-bm}(i) \cite{nagel2020adaround}. We include them, with their sources, to keep the paper self-contained. The statements specific to state trees are mainly \cref{prop:a-reach,prop:a-u2s1}, \cref{thm:a-forest} and \cref{thm:a-bm}(ii).

\subsection{Definitions}

\begin{definition}[State tree]\label{def:a-tree}
A \emph{state tree} $\T$ over $I$ consists of (a)~a rooted tree whose leaves are in bijection with $I$; an internal node may have a single child, so that one group of entries can carry several decisions in sequence; (b)~for every node $v$, a finite alphabet $S_v$ and a value map $c_v$ that assigns a real value to each state of $v$, possibly depending on the states of the ancestors of $v$ (a node whose value depends on the states of its ancestors is a \emph{lookup}); (c)~row scales $\rho\in\R^m$ and column bases $\beta\in\R^n$, stored in fixed formats (for example FP16). For an assignment $s=(s_v)_v$ of states, the represented matrix is $\What_{ij}(s)=\rho_i\beta_j\prod_{v\in\pi(i,j)}c_v(s)$, where $\pi(i,j)$ is the path from the root to the leaf of $(i,j)$. The set of matrices representable by $\T$ is $\M(\T)=\{\What(s)\}$. The leaf sets of the subtrees of $\T$ form a laminar family of groups of entries, the \emph{hierarchy} of $\T$. The \emph{depth} of $\T$ is the largest number of edges on a path from the root to a leaf, and the \emph{height} $h(v)$ of a node is the largest number of edges from $v$ down to a leaf of its subtree.
\end{definition}

The definition is deliberately broad: because a value map may depend on the states of the ancestors, any code that is decoded from the root down can be written as a state tree, so \cref{thm:a-formats} is a statement about costs rather than about expressiveness. Some results use the narrower \emph{core} model, in which no value map depends on the ancestors: \cref{prop:a-reach,prop:a-u2s1} are stated for it, and \cref{thm:a-bm}(ii) for node values that are affine in one bit or linear in one one-hot group. Trellis codes, whose overlapping windows share bits without nesting, are state trees only as a codebook over a whole block.

\begin{definition}[State-tree graph]\label{def:a-graph}
State-tree expressions are generated by $\mathsf E::=\T\mid\mathsf E+\mathsf E\mid\mathsf E\cdot\mathsf E\mid \mathrm{ref}(\mathsf E)$: a state tree, a \emph{sum node} (entrywise sum of operands of equal shape), a \emph{product node} (matrix product of compatible operands) and a \emph{reference} to an expression defined elsewhere. An expression in which some part is referenced more than once is a \emph{state-tree graph}, and the referenced part is a \emph{shared subtree}. The tables read by lookups are shared subtrees.
\end{definition}

\begin{definition}[Costs]\label{def:a-cost}
Under an assignment $s$, a node is \emph{inactive} if one of its ancestors has value $0$ or is inactive, or if its own value map is identically zero given the states of its ancestors; otherwise it is \emph{active}. The \emph{storage cost} is $\bits(s)=\sum_{v\ \text{active}}\log_2|S_v|$, with each shared subtree counted once, plus the cost of $\rho$ and $\beta$ in their formats. A decoder that reads the states from the root down knows which nodes are active, so this is the length of a decodable code, up to rounding that joint coding of several states makes negligible. The shape of the tree, including the order of rows and columns, is fixed per matrix shape and is not counted; a data-dependent order of the rows costs at most $\log_2 m!$ further bits, and can be folded into the neighbouring layer when every matrix that reads or writes the same dimension uses the same order, as all matrices that read or write the residual stream of a transformer must, including the embedding and the output head. Folding gives one order per dimension, shared by every matrix that reads or writes it (the gate and up projections of a gated MLP, for instance, share theirs); a separate order for each matrix instead costs $\log_2 m!/(mn)<\log_2 m/n$ bits per weight, which is negligible for large layers. If every entry is a sum of $t_{ij}$ elements of $\Pset$ times $\rho_i\beta_j$, the \emph{circuit cost} is the number of two-input additions used to evaluate $x\mapsto\What x$ by \cref{eq:circuit}, which is at most $\sum_i\big(\sum_jt_{ij}-1\big)^+$; sharing partial sums between outputs can reduce it \cite{voronenko2007mcm}.
\end{definition}

\subsection{What state trees express}

\begin{proposition}[Existing formats; \cref{prop:formats}]\label{thm:a-formats}
Each of the following is a state tree with at most five levels of decisions on any path (at most three for (i)--(v), except four for HiFloat4) whose storage cost is at most that of the format, and equal to it for (i)--(v): (i)~binary floating-point formats, including subnormals; (ii)~$b$-bit integers with a scale per group, and with a scale and a zero point per group; (iii)~block floating point and microscaling formats, including shared microexponents (MX6, MX9) \cite{rouhani2023sharedmx} and HiFloat4 \cite{luo2026hif4}; (iv)~ternary weights with a scale; (v)~codebook formats that store, for every block of $d$ entries, an index into a table of $d$-dimensional vectors; (vi)~$N{:}M$ sparsity, with the kept values in any of the formats above.
\end{proposition}

\begin{proof}
(i)~Give each entry a chain of three nodes: a sign node with values $\pm1$; an exponent node with values $2^{e-e_0}$ (and $2^{1-e_0}$ for the subnormal exponent code); and a mantissa leaf that is a lookup on the exponent, with value $1+m2^{-p}$ for normal exponent codes and $m2^{-p}$ for the subnormal code. Codes reserved for infinities and NaN do not occur as weights, but stay in the alphabet, so the exponent node costs the full exponent field. (ii)~A group node whose alphabet is the scale format, above leaves $\{0,\dots,2^b-1\}$ or a symmetric integer range. For a zero point, insert below the scale node a node of value $1$ whose state is the zero point $z$, and make each leaf a lookup with value $q-z$. (iii)~A block node with values $2^{e}$ (or the block scale) above leaves in the element format; shared microexponents and HiFloat4 insert one or two levels of one-bit shift nodes, one node per sub-block, between the block node and the leaves. (iv)~One node holding the scale above leaves $\{-1,0,1\}$. (v)~A block node whose state $k$ indexes the table; each of its $d$ leaves has a single state and is a lookup that returns its coordinate of the $k$-th table vector; the table is a shared subtree. (vi)~A group node over $M$ entries whose state is one of the $\binom MN$ admissible masks; the leaves outside the mask have value $0$ and are inactive, and the others carry the value format; the mask costs $\log_2\binom MN$ bits, no more than the mask field of the format. In cases (i)--(v) the stored states are exactly the fields of the format.
\end{proof}

\begin{proposition}[Refinement and sharing]\label{prop:a-refine}
(i)~If $\T'$ is obtained from $\T$ by inserting nodes whose alphabets contain a state of value $1$ (independent of the ancestors), or by enlarging alphabets, then $\M(\T)\subseteq\M(\T')$. (ii)~A tree is contained in any sum node that has it as a summand, provided the other summands can represent $0$. (iii)~Replacing several copies of a subtree by references to one shared subtree yields a subset of the original set of matrices, and saves the storage of all copies but one.
\end{proposition}

\begin{proof}
(i)~Give the inserted nodes their value-one state and copy all other states; the represented matrix is unchanged. (ii)~Set the other summands to zero. (iii)~A shared assignment is an unshared assignment in which all copies have equal states.
\end{proof}

\begin{theorem}[Counting; \cref{prop:complete}]\label{thm:a-count}
Let $Q\subset\R$ have $K$ elements. (i)~The flat tree (a root of value $1$ above leaves with alphabet $Q$, and $\rho=\beta=\mathbf1$) represents every matrix in $Q^{m\times n}$ with $mn\log_2K$ bits. (ii)~Any encoding, of fixed or variable length, whose codewords have at most $B$ bits represents at most $2^{B+1}-1$ matrices. Representing all of $Q^{m\times n}$ therefore needs $B\ge mn\log_2K-1$, and $B\ge mn\log_2K$ for fixed-length codes.
\end{theorem}

\begin{proof}
(i)~is immediate from the definition. (ii)~There are $\sum_{\ell=0}^{B}2^\ell=2^{B+1}-1$ binary strings of length at most $B$, and $|Q^{m\times n}|=K^{mn}=2^{mn\log_2K}$.
\end{proof}

\begin{theorem}[Density at the optimal price; \cref{prop:complete}]\label{thm:a-density}
Let $\varepsilon\in(0,1]$ and $G_\varepsilon=\{-1+(2k-1)\varepsilon:\ k=1,\dots,\lceil1/\varepsilon\rceil\}$. (i)~For every $W$ with $\|W\|_{\max}\le1$, the flat tree with leaf alphabet $G_\varepsilon$ achieves $\|W-\What\|_{\max}\le\varepsilon$ with $mn\log_2\lceil1/\varepsilon\rceil$ bits. (ii)~Any encoding whose codewords have at most $B$ bits and which achieves maximum error $\varepsilon$ on all such $W$ has $B\ge mn\log_2(1/\varepsilon)-1$. (iii)~The gap between (i) and (ii) is at most $mn\log_2(1+\varepsilon)+1$ bits. (iv)~For a general $W$, a root node with a power-of-two value $2^e$, where $2^{e-1}<\|W\|_{\max}\le2^e$, gives maximum error at most $2\varepsilon\|W\|_{\max}$, at the additional cost of storing $e$.
\end{theorem}

\begin{proof}
(i)~Consecutive points of $G_\varepsilon$ are $2\varepsilon$ apart, the first lies at $-1+\varepsilon$ and the last at least at $1-\varepsilon$, so every $u\in[-1,1]$ is within $\varepsilon$ of $G_\varepsilon$. (ii)~The max-norm balls of radius $\varepsilon$ around the codewords cover $[-1,1]^{mn}$; each has volume $(2\varepsilon)^{mn}$, so there are at least $\varepsilon^{-mn}$ codewords, and the bound follows as in \cref{thm:a-count}. (iii)~$\log_2\lceil1/\varepsilon\rceil\le \log_2(1/\varepsilon+1)=\log_2(1/\varepsilon)+\log_2(1+\varepsilon)$. (iv)~Apply (i) to $2^{-e}W$.
\end{proof}

\begin{theorem}[Sums of shift trees; \cref{prop:shift}]\label{thm:a-shift}
(i)~For every real $x\ne0$ and $t\ge1$ there are $p_1,\dots,p_t\in\Pset$ with $|x-\sum_\tau p_\tau|\le3^{-t}|x|$. (ii)~Every integer $z$ with $|z|<2^b$ is a sum of at most $\lceil(b+1)/2\rceil$ nonzero elements of $\Pset$, and of $b/3+O(1)$ on average over uniformly random $z$ \cite{reitwiesner1960}. (iii)~Hence for every $W$ and $t\ge1$ there is a sum of $t$ flat trees with values in $\Pset$ (the finitely many terms that occur form their alphabets), $\What=\diag(\rho)\big(\sum_{\tau=1}^tP_\tau\big)\diag(\beta)$, such that $|W_{ij}-\What_{ij}|\le3^{-t}|W_{ij}|$ for all entries; every matrix $\diag(\rho)Z\diag(\beta)$ with integer entries $|Z_{ij}|<2^b$ is represented exactly with $t=\lceil(b+1)/2\rceil$; and $x\mapsto\What x$ is computed with shifts, at most $\sum_i(\sum_jt_{ij}-1)^+$ additions, where $t_{ij}\le t$ is the number of nonzero terms of entry $(i,j)$, and $m+n$ multiplications.
\end{theorem}

\begin{proof}
(i)~If $|x|\in[2^k,2^{k+1}]$, the nearer of $2^k$ and $2^{k+1}$, with the sign of $x$, is within $2^{k-1}$ of $x$, and the distance relative to $|x|$ is largest, $\frac13$, at $|x|=3\cdot2^{k-1}$. So one term leaves a residual of at most $|x|/3$; repeat on the residual, and use the term $0$ once the residual vanishes. (ii)~The non-adjacent form of $z$ has at most $b+1$ signed binary digits, no two adjacent ones nonzero, and its average density of nonzero digits is $\frac13$ \cite{reitwiesner1960}. (iii)~Apply (i) or (ii) to every entry of $\diag(\rho)^{-1}W\diag(\beta)^{-1}$ ($\rho=\beta=\mathbf1$ is allowed); zero entries take the term $0$. In the product, every term is a shifted and signed copy of $\beta_jx_j$ added into the accumulator of $y_i$.
\end{proof}

\begin{proposition}[Reach of shared shifts; \cref{prop:reach}]\label{prop:a-reach}
Consider one column of a perfect binary tree of depth $D$, with a shift bit $b_e\in\{0,1\}$ (multiplier $2^{-b_e}$) on every edge and leaves in $\{0,\pm1\}$. For a leaf $\ell$ let $k_\ell=\sum_{e\in\pi(\ell)}b_e$, and let $L$ be the set of nonzero leaves. Integer exponents $(k_\ell)_{\ell\in L}$ are realizable if and only if $0\le k_\ell\le D$ for every $\ell\in L$ and, for every node $v$ with subtree $T_v$ and height $h(v)$,
\[
  \max_{\ell\in T_v\cap L}k_\ell-\min_{\ell\in T_v\cap L}k_\ell\;\le\;h(v).
\]
\end{proposition}

\begin{proof}
Let $p_v$ be the number of active bits on the path from the root to $v$. \emph{Necessity:} every leaf $\ell$ below $v$ has $p_v\le k_\ell\le p_v+h(v)$. \emph{Sufficiency:} choose the $p_v$ from the top, keeping $\max_{T_v\cap L}k-h(v)\le p_v\le\min_{T_v\cap L}k$ (no condition if $T_v\cap L$ is empty); this holds at the root with $p=0$ because $0\le k_\ell\le D$. For a child $c$ of $v$ with $T_c\cap L\neq\emptyset$, the admissible interval $[\max_{T_c\cap L}k-h(c),\ \min_{T_c\cap L}k]$ is nonempty by the condition at $c$; its left end is at most $\max_{T_v\cap L}k-h(v)+1\le p_v+1$ and its right end at least $\min_{T_v\cap L}k\ge p_v$. Hence it contains $p_v$ or $p_v+1$; take that value as $p_c$ and set $b_{(v,c)}=p_c-p_v$. At a leaf the interval is the single point $k_\ell$. Bits on edges into subtrees without nonzero leaves are arbitrary.
\end{proof}

\begin{proposition}[Two equivalent encodings; \cref{prop:reach}]\label{prop:a-u2s1}
For one column of a perfect binary tree with $m$ leaves, the encoding of \cref{prop:a-reach} (edge bits, leaves $\{0,\pm1\}$) and the encoding with one shift bit per internal node and leaves $\{0,\pm1,\pm\frac12\}$ represent the same set of columns, up to a factor $\frac12$ per column that can be absorbed into $\beta_j$. The first stores $2(m-1)+m\log_23$ bits and the second $(m-1)+m\log_25$, about $3.58$ and $3.32$ bits per weight for large $m$, with leaves coded jointly.
\end{proposition}

\begin{proof}
A bit on the edge into an internal node $c$ multiplies exactly the subtree of $c$, as the node bit of $c$ does; a bit on the edge into a leaf multiplies only that leaf, which is the leaf magnitude $\frac12$. This maps every edge-bit assignment to a node-bit assignment with root bit $0$ that represents the same column; conversely, a node-bit assignment with root bit $1$ is one with root bit $0$ times $\frac12$. The map is onto but not one-to-one, because a zero leaf ignores the bit on its edge.
\end{proof}

\begin{theorem}[Rank; \cref{sec:express}]\label{thm:a-rank}
(i)~A sum node of $r$ product nodes $u_kv_k^\top$ represents exactly the matrices of rank at most $r$. (ii)~Let $H\succ0$ with $H=LL^\top$. The minimizer of $f_H$ over matrices of rank at most $r$ is $\What=[WL]_rL^{-1}$, where $[\cdot]_r$ keeps the $r$ largest singular values. (iii)~Let also $G\succ0$ with $G=L_GL_G^\top$. The minimizer of $\tr\big(G(W-\What)H(W-\What)^\top\big)$ over matrices of rank at most $r$ is $\What=L_G^{-\top}[L_G^\top WL]_rL^{-1}$.
\end{theorem}

\begin{proof}
(i)~is the rank decomposition. (iii)~The objective equals $\|L_G^\top WL-L_G^\top\What L\|_F^2$, and $\What\mapsto L_G^\top\What L$ is a linear bijection that preserves rank, so the Eckart--Young theorem \cite{eckart1936} gives the truncated singular value decomposition. (ii)~is the case $G=I$. Singular $H$ or $G$ are handled by damping.
\end{proof}

\begin{proposition}[Shared subtrees; \cref{sec:express}]\label{prop:a-share}
Let $W_1,\dots,W_g$ have $n$ columns and input statistics $H_1,\dots,H_g\succeq0$, and write $W_i\approx C_iD$ with one shared $D\in\R^{r\times n}$. (i)~Exact representation is possible if and only if $r\ge\rank[W_1;\dots;W_g]$. (ii)~For fixed $C_i$, the minimizers of $\sum_if_{H_i}(C_iD)$ are the solutions of the generalized Sylvester equation $\sum_i(C_i^\top C_i)\,D\,H_i=\sum_iC_i^\top W_iH_i$. (iii)~For $g=2$, write this equation as $ADB+CDE=K$. If $A,B\succ0$ (for instance $C_1$ of full column rank and $H_1\succ0$), let $P^\top AP=I$, $P^\top CP=\diag(\lambda)$, $Q^\top BQ=I$ and $Q^\top EQ=\diag(\sigma)$ be the generalized eigendecompositions. Then $D=P\big[(P^\top KQ)\oslash(\mathbf1\mathbf1^\top+\lambda\sigma^\top)\big]Q^\top$, with $\oslash$ the entrywise division.
\end{proposition}

\begin{proof}
(i)~$W_i=C_iD$ for all $i$ if and only if every row of every $W_i$ lies in the row space of $D$. (ii)~The objective is a convex quadratic function of $D$; set its gradient to zero. (iii)~Substitute $D=PXQ^\top$ and multiply by $P^\top$ on the left and by $Q$ on the right to obtain $X+\diag(\lambda)X\diag(\sigma)=P^\top KQ$, which is solved entrywise; the denominators are at least $1$ because $\lambda,\sigma\ge0$, as $C$ and $E$ are positive semidefinite.
\end{proof}

\subsection{Compression, circuits and Boltzmann machines}

\begin{theorem}[One optimizer; \cref{prop:mckp}]\label{thm:a-mckp}
Let every matrix $\mu$ have a finite menu $\mathcal O_\mu$ of options with sizes $b_\mu(o)$ and accuracy costs $c_\mu(o)$, and consider $\min\sum_\mu c_\mu(o_\mu)$ subject to $\sum_\mu b_\mu(o_\mu)\le B$. (i)~Replacing every menu by its Pareto set, the options not dominated in both size and cost, does not change the optimum. (ii)~Dynamic programming over a grid of $U$ size units solves the problem exactly, with sizes rounded up to the grid so that its solutions are feasible, in $O\big(U\sum_\mu|\Pareto(\mathcal O_\mu)|\big)$ time \cite{sinha1979mckp}; its value lies between the optima of the unrounded problem for the budget $B$ and for $B$ reduced by one grid unit per matrix. (iii)~The optimum is at most the cost of any single method or uniform setting that fits the budget.
\end{theorem}

\begin{proof}
(i)~Replacing an option by one that dominates it keeps the budget and does not increase the cost. (ii)~The recursion $\mathrm{DP}_\mu(u)=\min_o\{\mathrm{DP}_{\mu-1}(u-\lceil b_\mu(o)\rceil)+c_\mu(o)\}$ enumerates all choices. Its solutions are feasible for $B$, and a choice that fits $B$ reduced by one unit per matrix still fits after rounding up, since each size grows by less than one unit. (iii)~One method applied to every matrix is one feasible choice.
\end{proof}

\begin{theorem}[Shared subtrees in the optimizer; \cref{prop:mckp}]\label{thm:a-forest}
Let the candidate groups for shared subtrees form a forest whose leaves are single matrices, and let every internal node $v$ offer a set $\mathcal S_v$ of (size, cost) options, each sharing one subtree among all matrices below $v$. With additive costs, the recursion $\Pareto(v)=\Pareto\big((\bigoplus_{c\in\mathrm{ch}(v)}\Pareto(c))\cup\mathcal S_v\big)$, with $\oplus$ the Minkowski sum of (size, cost) sets and $\Pareto(\mu)=\Pareto(\mathcal O_\mu)$ at the leaves, followed by \cref{thm:a-mckp} over the roots, is exact over all configurations in which the shared groups are pairwise disjoint.
\end{theorem}

\begin{proof}
In an admissible configuration, the matrices below $v$ are either covered by one option at $v$ or split among the children of $v$ with independent configurations whose sizes and costs add. By induction, the attainable pairs below $v$ are $(\bigoplus_c\mathcal C_c)\cup\mathcal S_v$, where $\mathcal C_c$ is the set of attainable pairs below $c$, and by \cref{thm:a-mckp}(i) it suffices to keep Pareto sets at every step. With sizes rounded up to the grid of \cref{thm:a-mckp} once, in the menus, all sums stay on the grid and every Pareto set has at most $U+1$ points.
\end{proof}

\begin{proposition}[Limits of shift-and-add circuits; \cref{sec:circuits}]\label{prop:a-lower}
(i)~Consider straight-line programs whose instructions add or subtract two earlier values (inputs or results), each shifted by at most $S$ positions. If such a program of length $L$ computes $x\mapsto Wx$ exactly, then for all but a vanishing fraction of $W\in\{0,\dots,2^b-1\}^{N\times N}$, $L\log(NLS)=\Omega(N^2b)$; for $S$ polynomial in $N$ and $b$ this gives $L=\Omega\big(N^2b/\log(Nb)\big)$. (ii)~$O(N^2b/\log N)$ additions suffice: split $W$ into $b$ binary matrices, group $g\approx\log_2N-\log_2\log_2N$ rows, add every input once into the bucket indexed by its $g$-bit column pattern, combine the bucket sums into the outputs of the group, and combine the bit planes with shifts \cite{arlazarov1970,liberty2009mailman}.
\end{proposition}

\begin{proof}
(i)~An instruction is described by two operand indices among at most $N+L$ values, two shifts and a sign, i.e.\ by $O(\log(NLS))$ bits, and the program also names its $N$ outputs. So there are $2^{O(L\log(NLS))+N\log(N+L)}$ programs, while there are $2^{N^2b}$ matrices, and different matrices need different programs, since a program determines the map it computes. If $L\ge N^2b$ the second bound holds trivially; otherwise $\log(NLS)=O(\log(Nb))$ for $S$ polynomial in $N$ and $b$. (ii)~For each of the $b$ bit planes and each of the $N/g$ row groups, adding the $N$ inputs into $2^g$ buckets takes $N$ additions, and each of the $g$ outputs sums at most $2^{g-1}$ buckets, so the group costs at most $N+g2^{g}\le2N$ additions by the choice of $g$. In total this is $O(N^2b/g)=O(N^2b/\log N)$, plus $O(Nb)$ additions for the bit planes.
\end{proof}

\begin{theorem}[The Boltzmann machine of a tree; \cref{prop:bm}]\label{thm:a-bm}
Fix $\rho$ and $\beta$, and encode the states of a state tree for $W$ with binary units: a leaf with values $a_1,\dots,a_K$ as a one-hot group $z_{ij,1},\dots,z_{ij,K}$; a shift bit with multiplier $r^b$ as the affine factor $1+(r-1)b$; a gate as a bit that multiplies its subtree; and an index into a fixed table as a one-hot group. As the energy take $f_H$ of \cref{eq:fH}, or $f_{G,H}=\tr(GEHE^\top)$, plus a penalty $\lambda(\sum_kz_k-1)^2$ for every one-hot group. (i)~With the node states fixed, the energy is quadratic in the leaf units. Write $\What_{ij}=\sum_kv_{ij,k}z_{ij,k}$ with $v_{ij,k}=\rho_i\beta_j\pi_{ij}a_k$, where $\pi_{ij}$ is the product of the fixed node values on the path. Under $f_H$, units of entries $(i,j)$ and $(i,l)$ with $j\ne l$ are coupled with $J=2v_{ij,k}v_{il,k'}H_{jl}$, the coefficient of $z_{ij,k}z_{il,k'}$ in \cref{eq:bm}, and units in different rows are not coupled. Under $f_{G,H}$, units of entries $(i,j)\ne(i',l)$ are coupled with $J=2v_{ij,k}v_{i'l,k'}G_{ii'}H_{jl}$. Two units of the same leaf ($k\ne k'$) are coupled with $J=2v_{ij,k}v_{ij,k'}H_{jj}+2\lambda$, with $G_{ii}H_{jj}$ in place of $H_{jj}$ under $f_{G,H}$. Every unit has the field $h_{ij,k}=v_{ij,k}^2H_{jj}-2v_{ij,k}(WH)_{ij}-\lambda$, or $v_{ij,k}^2G_{ii}H_{jj}-2v_{ij,k}(GWH)_{ij}-\lambda$ under $f_{G,H}$, up to a constant. (ii)~With the node states free, and every node value affine in one bit or linear in one one-hot group, the energy is a polynomial of degree at most $2D+2$ for a tree of depth $D$, and each of its terms involves the units of at most two root-to-leaf paths. Auxiliary units and penalties reduce it to a quadratic function \cite{boros2002pseudo,ishikawa2011transformation}. (iii)~If $\lambda$ exceeds the range of the error over all binary assignments, every ground state satisfies the one-hot constraints, and the matrix it decodes to minimizes the error over $\M(\T)$.
\end{theorem}

\begin{proof}
(i)~With $e_i$ the $i$-th row of $E$, $f_H=\sum_ie_iHe_i^\top$ and $f_{G,H}=\sum_{i,i'}G_{ii'}e_iHe_{i'}^\top$. Substitute $\What_{ij}$, which is linear in the leaf units, and use $z^2=z$. (ii)~By assumption, each node value is affine in one bit or linear in a one-hot group, so $\What_{ij}$ is a polynomial of degree at most $D+1$ in the units on its path, and the error is quadratic in the entries. A lookup whose value depends on its own state and on the state of an ancestor is bilinear in their units, and raises the degree by at most one for each ancestor it reads. (iii)~An assignment that violates a one-hot constraint pays a penalty of at least $\lambda$, which then exceeds the difference between the error of any feasible assignment and the smallest error overall.
\end{proof}

\paragraph{Convolutions.} A convolution computes $W\,\mathrm{patch}(p)$ at every position $p$ with the same kernel matrix
$W$. Written as one matrix acting on the whole input, its blocks for different positions are copies of $W$ at shifted
columns, and each copy is a reference to one shared subtree. Grouped and depthwise convolutions are block-diagonal over
channel groups, that is, disjoint channel subtrees.

\subsection{Experimental details}\label{app:exp}

\paragraph{Common setup.} Perplexity is measured at context 2048 on WikiText-2, C4 and FineWeb; FP16 LLaMA-3-8B gives 6.14, 9.44 and 7.85 in the same pipeline. Matrices are compressed one block after another, with Hessians computed afresh from 128 FineWeb calibration sequences and keeping the attention-sink token. MMLU and GSM8K are evaluated 5-shot with lm-eval, with the same settings for every model. Compression and healing use PyTorch 2.7.1 and Transformers 5; serving uses vLLM 0.32.0 with PyTorch 2.13.0 and CUDA 12.9. All GPU measurements are on an A100-SXM4-40GB.

\paragraph{Lossless storage.} Each row is split into groups of 64 consecutive weights. The group node stores the largest BF16 exponent $E$ of the group (8 bits); each weight stores the state $D=E-e\in\{0,\dots,14\}$ of its exponent $e$ (4 bits) and a leaf with its sign and 7-bit mantissa (8 bits), 12.125 bits per weight. The state 15 is an escape: a weight with $D\ge15$, which includes exact zeros and subnormals, has its leaf set to zero and is stored in a side list with its 16 BF16 bits and its row and column, 80 bits in all. Escapes are rare, 0.0095\% of the weights of Llama-3.2-1B and 0.0097\% of those of Qwen3.5-9B and at most 0.014\% in any layer, so whole models, embeddings and output head included, take 12.133 bits per weight on disk. At runtime every escape is repacked as one 32-bit entry, and the kernel layout of a single LLaMA-3-8B matrix takes 12.137 bits per weight. The fused kernel patches escaped weights in registers from a 64-byte block per work unit, holding a bit mask, a count and up to 13 entries, and reads any further escapes from a global list. Every weight of every matrix tested decodes to its exact BF16 bits.

\paragraph{Whole-model optimizer.} Every block linear layer receives the same menu of quantization, pruning and low-rank edits, alone and combined, with no rotation and no trellis-coded quantizer; the menus will be described in a follow-up paper. The cost of an option is the measured curvature cost of \cref{sec:directions}, on 16 calibration sequences. GPTQ at 4, 3 and 2 bits is measured for all 224 matrices; the costs of the other options are first interpolated along each matrix's curve of cost against $f_H$. Every option the solver picks is then built and measured, and the knapsack is solved again until the whole plan has been measured; in the 4.25-bit run (26.6\% of the FP16 bytes) this took nine rounds, and the 3.25-bit run of \cref{sec:directions} uses 20.3\% of the bytes of the block weights. Three further rounds re-measure the costs with the whole plan applied.

\paragraph{Ising healing.} The 2.44-bit model, an illustration, is Qwen2.5-1.5B-Instruct compressed without training by the optimizer with a wider menu, to 15.2\% of the bytes of its block weights; the menu will be described in a follow-up paper. Healing starts from a rebuild that reruns the whole pipeline with a recorder of the leaf states. Its re-measured costs give a slightly different plan, with perplexities of 13.74, 20.95 and 20.88 on WikiText-2, C4 and FineWeb against 13.63, 20.75 and 20.71 for the first run; a second rebuild gave 13.72, 20.84 and 20.85, so runs differ by about 0.1. All gains are measured against the rebuild. Only leaves of the GPTQ family, 6.8\% of the weights, can move; all other leaves stay fixed. Each step estimates the gradient of the KL divergence and its Kronecker factors $H=\mathbb E[xx^\top]$ and $G=\mathbb E[gg^\top]$ on about 45k tokens, with labels sampled from the student. A move must lower the energy by more than a threshold, set so that a fraction $q$ of the weights qualifies; the lowest energy among no move, the greedy update and simulated bifurcation started from it is kept; and the step is accepted only if the validation KL improves, after which $q$ grows by a factor of 1.5, or else shrinks by a factor of 0.3. There are 40 steps. After healing, the perplexities are 12.85, 19.51 and 20.02. In a solver ablation with the same model, data and number of steps, greedy updates alone reach a validation KL of 0.378 (GSM8K 20.32, MMLU 49.49), simulated bifurcation 0.371 (21.00, 49.29), local search 0.369 and random moves 0.383; the reported run keeps the best of no move, the greedy update and simulated bifurcation at every step.

\paragraph{Image coding on CLIC.} On the 41 images of the CLIC 2020 professional validation set, the GPU codec needs 8.51 bits per pixel, against 7.88 for JPEG XL at effort 9 (7.97 at effort 7), 8.81 for lossless WebP and 11.71 for PNG, all run through imagecodecs 2026.3.6 on CPU (libjxl 0.11.2; libwebp 1.6.0 at method 4, against method 6 on Kodak; libpng 1.6.55 with zlib at level 9). The whole set decodes at 2.86 Gpixel/s end to end, 3.46 Gpixel/s for the kernels alone, and one image at a time at 0.59--0.66 Gpixel/s. JPEG 2000 and HTJ2K were measured on Kodak only.


\end{document}